\documentclass{article}

\usepackage{microtype}
\usepackage{graphicx}
\usepackage{booktabs} %
\usepackage{pgfplots}
\usepackage{pgfplotstable,pgf}
\pgfplotsset{compat=newest}
\usepackage{pdftexcmds}
\usepgfplotslibrary{fillbetween}

\usepackage{tikz}
\usetikzlibrary{positioning, shapes.geometric, arrows.meta, calc, backgrounds, shadows}

\definecolor{mainblue}{RGB}{70, 130, 180}
\definecolor{lightgray}{RGB}{245, 245, 245}
\definecolor{darkblue}{RGB}{25, 25, 112}

\tikzset{
    neuron/.style={circle, draw=darkblue!80, fill=white, inner sep=0pt, minimum size=14pt, thick, copy shadow={shadow xshift=1pt, shadow yshift=-1pt, fill=black!20}},
    box/.style={draw=darkblue, fill=mainblue!10, rounded corners=4pt, minimum width=1.4cm, minimum height=0.7cm, thick, font=\bfseries, copy shadow={shadow xshift=1.5pt, shadow yshift=-1.5pt, fill=black!15}},
    arrow/.style={-{Stealth[scale=1.2]}, thick, draw=black!70},
    label_style/.style={font=\footnotesize\sffamily\bfseries, color=black!60}
}

\definecolor{c0}{HTML}{1f77b4}
\definecolor{c1}{HTML}{ff7f0e}
\definecolor{c2}{HTML}{2ca02c} %
\definecolor{c3}{HTML}{d62728} %
\definecolor{c4}{HTML}{9467bd} 
\definecolor{c5}{HTML}{8c564b}
\definecolor{c6}{HTML}{e377c2}
\definecolor{c7}{HTML}{7f7f7f}
\definecolor{c8}{HTML}{bcbd22}
\definecolor{c9}{HTML}{17becf}

\usepackage[preprint]{icml2026}

\usepackage{amsmath}
\usepackage{amssymb}
\usepackage{mathtools}
\usepackage{amsthm}
\usetikzlibrary{positioning, chains, fit, shapes, arrows.meta}
\theoremstyle{plain}
\newtheorem{theorem}{Theorem}[section]

\theoremstyle{definition}

\theoremstyle{remark}

\usepackage[textsize=tiny]{todonotes}

\makeatletter
\AtBeginDocument{%
    \catcode`_=12
    \begingroup\lccode`~=`_
    \lowercase{\endgroup\let~}\sb
    \mathcode`_="8000
    \immediate\write\@auxout{\catcode`_=12 }%
    \immediate\write\@auxout{\catcode`^=12 }%
}
\makeatother

\usepackage[colorlinks]{hyperref}       %
\usepackage{hyperref}       %
\usepackage{colortbl}       %
\usepackage{amsfonts}       %
\usepackage{nicefrac}       %
\usepackage{microtype}      %
\usepackage{xcolor}         %
\usepackage{float}
\usepackage{placeins}
\usepackage{pifont}%
\newcommand{\cmark}{\ding{51}}%
\newcommand{\xmark}{\ding{55}}%
\usepackage{multirow}
\usepackage{capt-of}

\usepackage[american]{babel}
\usepackage{mathtools} %
\usepackage{booktabs} %
\usepackage{tikz} %
\usepackage{tcolorbox} %
\usepackage{bm}
\usepackage{dblfloatfix}

\usepackage{booktabs}       %
\usepackage{amsfonts}       %
\usepackage{enumitem}
\usepackage{amsmath}
\usepackage{subcaption}
\usepackage{adjustbox}
\usepackage{soul}
\usepackage{tcolorbox}
\tcbuselibrary{skins, breakable}

\DeclareMathOperator*{\argmin}{argmin}

	\definecolor{sthlmLightBlue}{RGB}{214,237,252} %
		
	\definecolor{sthlmBlue}{RGB}{0,110,191} %

	\definecolor{sthlmLightGreen}{RGB}{213,247,244} %
		
	\definecolor{sthlmGreen}{RGB}{0,134,127} %

	\definecolor{sthlmLightGrey}{RGB}{213,217,225} %
		
	\definecolor{sthlmGrey}{RGB}{245,243,238} %
		
	\definecolor{sthlmDarkGrey}{RGB}{51,51,51} %

	\definecolor{sthlmLightOrange}{RGB}{255,215,210} %
		
	\definecolor{sthlmOrange}{RGB}{221,74,44} %

	\definecolor{sthlmLightPurple}{RGB}{241,230,252} %
		
	\definecolor{sthlmPurple}{RGB}{93,35,125} %

	\definecolor{sthlmLightRed}{RGB}{254,222,237} %
		
	\definecolor{sthlmRed}{RGB}{196,0,100} %

	\definecolor{sthlmYellow}{RGB}{252,191,10} %

\definecolor{sthlmGrey}{RGB}{245,245,245}

\newtheoremstyle{contribution}%
  {12pt}%
  {12pt}%
  {\itshape}%
  {}%
  {\bfseries\color{sthlmBlue}}%
  {.}%
  {.5em}%
  {\thmname{#1}\thmnumber{ #2}\thmnote{ (#3)}} %

\newtheoremstyle{contribution}
  {12pt}{12pt} %
  {\itshape}    %
  {}            %
  {\bfseries\color{sthlmBlue}} %
  {.}           %
  {.5em}        %
  {\thmname{#1}\thmnumber{ #2}\thmnote{ (#3)}}

\theoremstyle{contribution}

\newtcolorbox[auto counter]{experimentbox}[2][]{
    enhanced,
    before skip=2mm,after skip=2mm,
    colback=white,
    colframe=sthlmBlue,
    boxrule=0.5mm,
    attach boxed title to top left={xshift=1mm,yshift=-2mm,yshifttext=-1mm},
    fonttitle=\bfseries,
    coltitle=white,
    colbacktitle=sthlmBlue,
    title={Experiment \thetcbcounter: {#2}},
    sharp corners,
    leftrule=2mm, %
    rightrule=0.2mm,
    bottomrule=0.2mm,
    toprule=0.2mm,
    boxed title style={sharp corners, size=small},
    #1
}

\hypersetup{
 linkcolor=sthlmBlue 
,citecolor=sthlmRed
,urlcolor=sthlmRed
}

\usepackage{colortbl}%
  
\usepackage{booktabs}

\def\approxprop{%
  \def\p{%
    \setbox0=\vbox{\hbox{$\propto$}}%
    \ht0=0.3ex \box0 }%
  \def\s{%
    \vbox{\hbox{$\sim$}}%
  }%
  \mathrel{\raisebox{0.3ex}{%
      \mbox{$\overset{\s}{\p}$}%
    }}%
}
\usepackage{cancel}

\icmltitlerunning{On the Non-Orthogonality of Aleatoric and Epistemic Uncertainty}
\icmltitlerunning{Measuring Orthogonality as the Blind-Spot of Uncertainty Disentanglement}

\author{
  Ivo Pascal de Jong\\ University of Groningen \\ Bernoulli Institute \And
  Andreea Ioana Sburlea\\ University of Groningen \\ Bernoulli Institute  \AND
  Matthia Sabatelli\\ University of Groningen \\ Bernoulli Institute  \And
  Matias Valdenegro-Toro \\ University of Groningen \\ Bernoulli Institute
}

\begin{document}

\twocolumn[
\icmltitle{Measuring Orthogonality as the Blind-Spot of Uncertainty Disentanglement}

\icmlsetsymbol{equal}{*}

\begin{icmlauthorlist}
\icmlauthor{Ivo Pascal de Jong}{yyy}
\icmlauthor{Andreea Ioana Sburlea}{yyy}
\icmlauthor{Matthia Sabatelli}{yyy}
\icmlauthor{Matias Valdenegro-Toro}{yyy}
\end{icmlauthorlist}

\icmlaffiliation{yyy}{Department of Artificial Intelligence, Bernoulli Instutute, University of Groningen, Groningen, The Netherlands}

\icmlcorrespondingauthor{Ivo Pascal de Jong}{ivo.de.jong@rug.nl}
\icmlkeywords{Uncertainty Quantification, Bayesian Neural Networks, Uncertainty Disentanglement, aleatoric uncertainty, epistemic uncertainty}

\vskip 0.3in
]

\printAffiliationsAndNotice{}  %

\begin{abstract}
    Aleatoric (data) and epistemic (knowledge) uncertainty are textbook components of Uncertainty Quantification. Jointly estimating these components has been shown to be problematic and non-trivial. As a result, there are multiple ways to disentangle these uncertainties, but current methods to evaluate them are insufficient. We propose that aleatoric and epistemic uncertainty estimates should be \textit{orthogonally} disentangled -- meaning that each uncertainty is not affected by the other -- a necessary condition that is often not met. We prove that orthogonality and consistency and necessary and sufficient criteria for disentanglement, and construct Uncertainty Disentanglement Error as a metric to measure these criteria, with further empirical evaluation showing that finetuned models give different orthogonality results than models trained from scratch and that UDE can be optimized for through dropout rate. We demonstrate a Deep Ensemble trained from scratch on ImageNet-1k with Information Theoretic disentangling achieves consistent and orthogonal estimates of epistemic uncertainty, but estimates of aleatoric uncertainty still fail on orthogonality.

\end{abstract}

\section{Introduction}\label{sec:introduction}

Bayesian Neural Networks (BNNs) quantify uncertainty in a component for aleatoric uncertainty, which stems from the data,  and a component for epistemic uncertainty, which arises from the model. This distinction enables decisions that depend on the underlying source of uncertainty: samples with high epistemic uncertainty (EU) can be deferred to a more reliable model or a human expert, while samples dominated by aleatoric uncertainty (AU) cannot get better predictions \citep{van2022certainty}. As a concrete example let us consider Table \ref{tab:orthogonal_uncertainty_concept}, where an unfamiliar but clearly visible object should result in high EU, and a better model or a human may be able to classify this object correctly. Meanwhile an occluded picture where the object is entirely not visible should have high AU, and the best way to get a better classification is to take a new picture. To be able to attribute uncertainty to either aleatoric or epistemic uncertainty, and therefore affect subsequent decision making, the estimates of aleatoric and epistemic uncertainty need to be \textit{orthogonally} disentangled. That is, a change in aleatoric uncertainty should not affect epistemic uncertainty, and vice versa. In this work, we will show that orthogonality is generally not achieved, and that standard methods of measuring the quality of disentanglement neglect orthogonality and are therefore insufficient.

\begin{table}[t]
     \begin{center}
           \caption{Example of when orthogonal disentanglement of aleatoric and epistemic uncertainty is necessary. With high EU, we could ask a human or another model to classify the image. When the uncertainty is due to AU, we know that the image cannot be classified reliably. Deferring an image with high AU to a human-in-the-loop does not lead to better outcomes, while under high EU it can.}      \label{tab:orthogonal_uncertainty_concept}
     \begin{tabular}{p{1cm}  p{2.5cm}  p{2.5cm}  }
     \toprule
        Object & \textbf{Unfamiliar} & \textbf{Not visible }\\ 
        \midrule
         & \includegraphics[width=\linewidth]{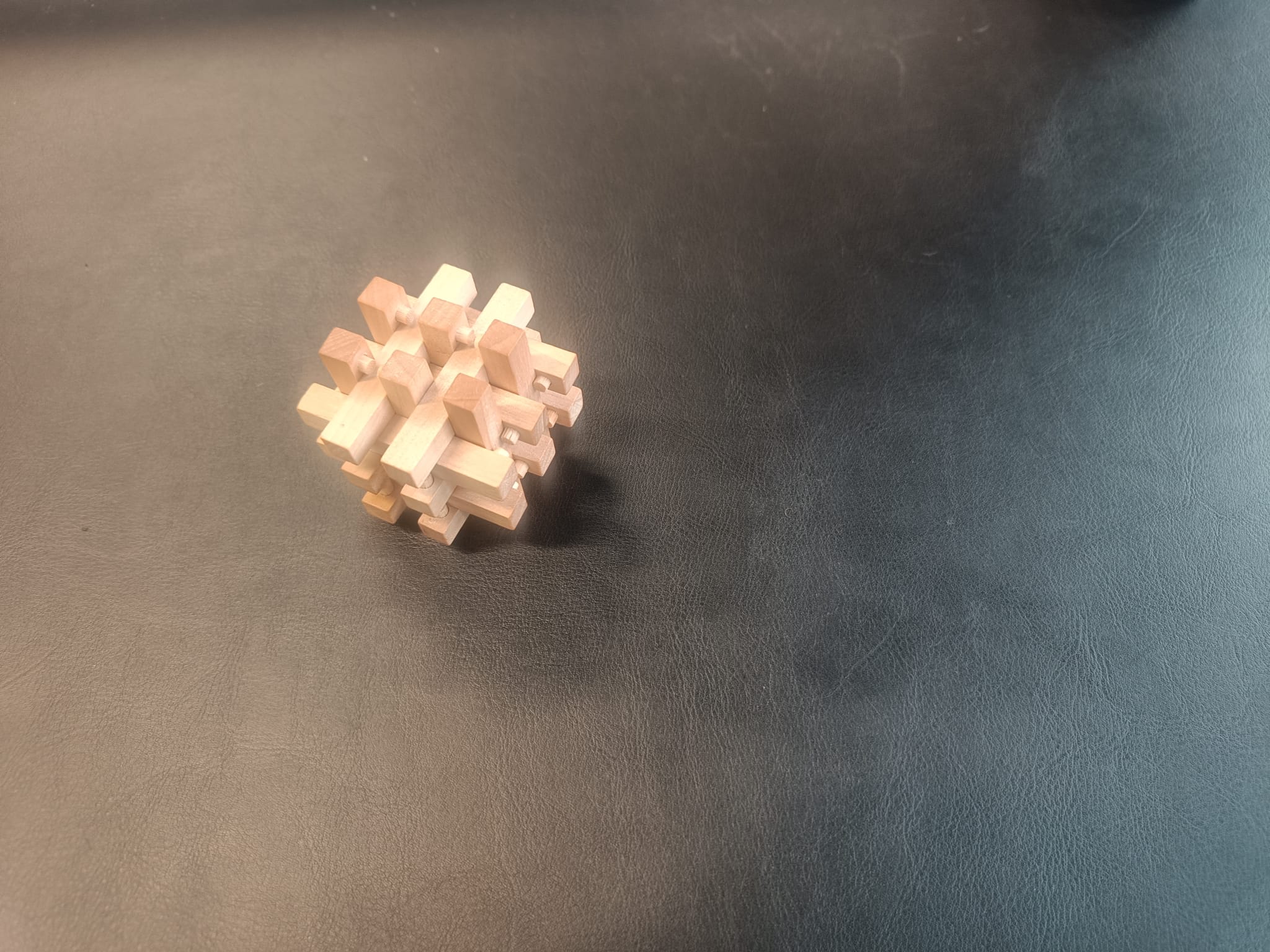} &  \includegraphics[width=\linewidth]{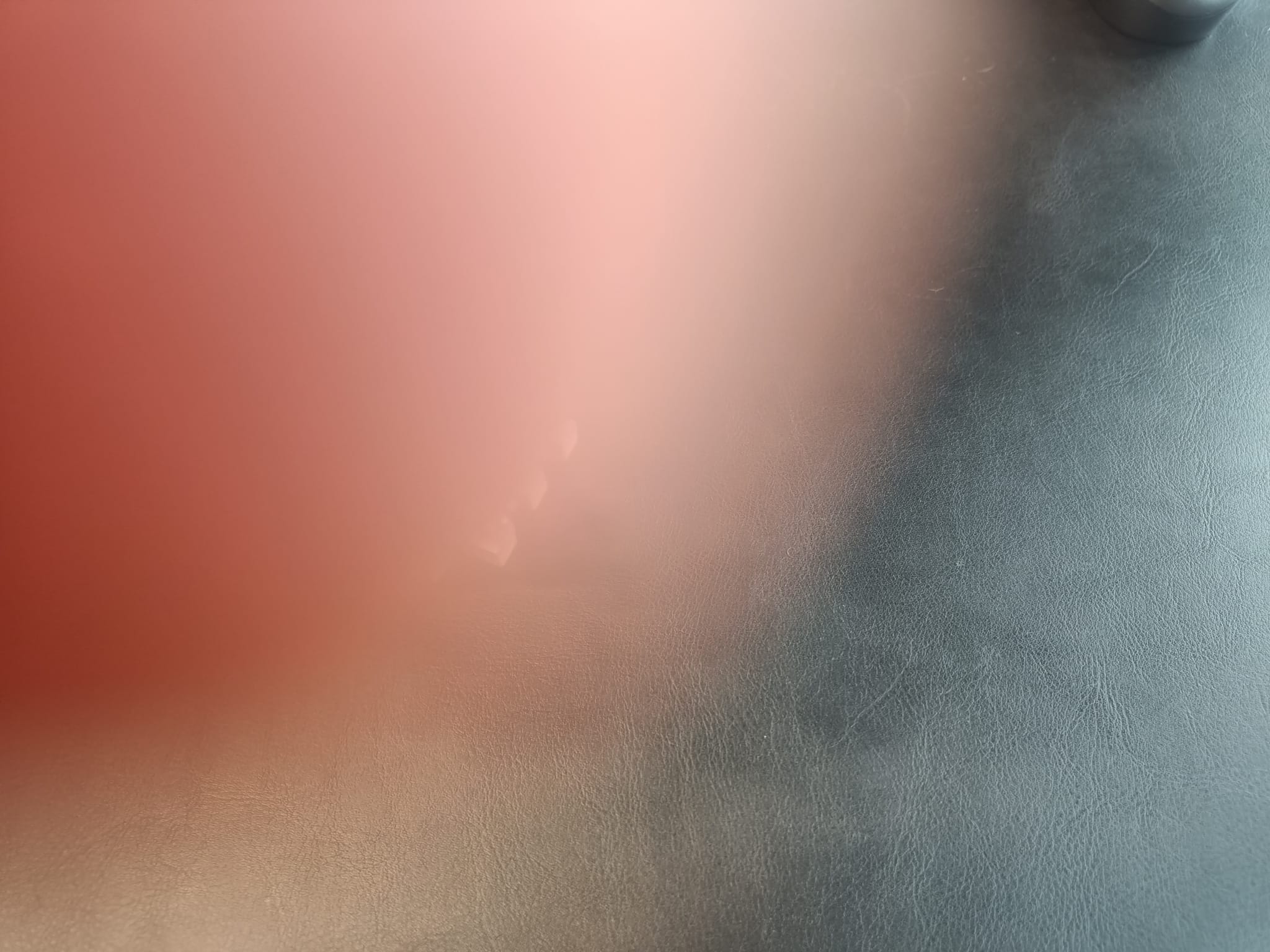} \\\midrule
        \textbf{Goal} & EU$\uparrow$ & AU $\uparrow$ \\\midrule
        \textbf{Solution} & Ask a human & Take new picture \\\midrule
        \textbf{Reality} & EU$\uparrow$ \& AU $\downarrow$ & AU$\uparrow$ \& EU $\uparrow$ \\
        \bottomrule
      \end{tabular}

      \end{center}
\end{table}

\paragraph{Different Disentangling Methods}

A popular approach to disentangling aleatoric and epistemic uncertainty is through Information Theoretic (IT) measures. However, this Information Theoretic approach has been challenged \citep{wimmer2023quantifying, mucsanyi2024benchmarking, kotelevskii2024predictive} as there is a major limitation inherent in the way these measures are formulated. In IT disentangling, total uncertainty $u_p$, epistemic uncertainty $u_e$ and aleatoric uncertainty $u_a$ all have the same (non-negative) scale and the same maximum value, but $u_p = u_a + u_e$. Therefore, any samples for which the model predicts maximum $u_a$ can only have $u_e=0$ \cite{wimmer2023quantifying}. We provide a demonstration of this phenomenon with toy data in Appendix \ref{sec:it_triangles}.

An alternative disentangling method exists \citep{kendall2017uncertainties, vranken2021uncertainty, van2022certainty} where $u_a$ is directly estimated as a variance in the logits, and $u_e$ is observed as the variance due to the model. We refer to this approach as \textit{Gaussian Logits}, since it assumes Gaussian distributed logits. %

Wanting to compare the estimates of $u_a$ and $u_e$ for IT disentangling and Gaussian Logits disentangling, we found that there were \textit{no existing methods to quantify the performance of orthogonal disentanglement of aleatoric and epistemic uncertainty}. Evaluation methods exist, but they do not consider orthogonality. We show that consistency and orthogonality are necessary and sufficient conditions for uncertainty disentanglement and propose a novel evaluation method to quantify compliance to these conditions, resulting in the \textit{Uncertainty Disentanglement Error} (UDE).

\paragraph{Contributions}

Previous work from \citet{valdenegro2022deeper, wimmer2023quantifying, mucsanyi2024benchmarking} has shown that there are problems with uncertainty disentanglement, but they do not measure the quality of orthogonal disentanglement. By combining two experiments that manipulate aleatoric and epistemic uncertainty orthogonally (that is, changing one while keeping the other the same), we can study whether the estimated aleatoric and epistemic uncertainty also behave orthogonally. Previous practice evaluates downstream task performance or correlation between aleatoric and epistemic estimates \citep{mucsanyi2024benchmarking}. Our evaluation is the first to evaluate whether disentanglement is orthogonal.

Our contributions are as follows:
\begin{itemize}[noitemsep, topsep=0pt]
    \item In Section \ref{sec:theory} we formally define orthogonal disentanglement, and prove that evaluating aleatoric and epistemic uncertainty in isolation is not sufficient, and that correlation between estimates does not constitute a failure of disentanglement. We argue that consistent and orthogonal disentanglement is necessary and sufficient to be able to distinguish between aleatoric and epistemic uncertainty. 
    \item In Section \ref{sec:experimental_setup} we provide a methodology to evaluate whether models comply with requirements of consistency and orthogonality based on well-established experiments \citep{wimmer2023quantifying, barandas2024evaluation}. We find that consistency is generally achieved, but orthogonality is not. %

    \item Based on the theoretical foundations of consistency and orthogonality, we introduce the Uncertainty Disentanglement Error (UDE) as a metric in Section \ref{sec:UDE}. We provide a \texttt{Python} package to calculate UDE for both \texttt{scikit-learn} and \texttt{PyTorch} models.\footnote{UDE Package: \url{https://anonymous.4open.science/r/disentanglement_error-8CEE}}. Using UDE we extensively compare different methods of disentangling uncertainties over multiple models and datasets. We show that Deep Ensembles with Information Theoretic disentanglement performs best, but that it does not achieve orthogonality. We evaluate UDE empirically and show that it is robust to changes in configuration, can be optimized for effectively, and that finetuned models give qualitatively different behavior than models trained from scratch.

    \item In Section \ref{sec:UDE} we also show that a Deep Ensemble with Information Theoretic disentangling trained on ImageNet-1k is able to estimate epistemic uncertainty consistently and orthogonally, but that estimates of aleatoric uncertainty are not orthogonal, adding nuance to previous claims that aleatoric and epistemic uncertainty are correlated \cite{mucsanyi2024benchmarking}.

\end{itemize}

\section{Background on Disentangling Methods}\label{sec:background}

Single point Neural Networks for classification typically aim to find the optimal parameters $\theta$ that minimise the empirical loss $\mathcal{L}(y_{pred}; y_{true})$ for some dataset $D = \{\mathbf{X}, \mathbf{y}\}$ such that $\theta = \argmin_\theta \mathcal{L}(f_\theta(x); y)$. For classification tasks this is extended with a Softmax activation function to predict the probability of some class $c$ so that we can do inference as $p(y\,{=}\,c\,|\,x, \theta)$, which accounts for AU. Bayesian Neural Networks expand this by considering all likely values of $\theta$ for the dataset $D$ to account for EU. Following \citet{malinin2018predictive}, the classification is then determined as 

\begin{equation}\label{eq:aleatoric_epistemic}
p(y|x) = \int \underbrace{p(y\,{=}\,c\,|\, x, \theta)}_{\text{Aleatoric}} \underbrace{p(\theta | D)}_{\text{Epistemic}} d\theta.
\end{equation}

MC-Dropout~\citep{gal2016dropout}, MC-DropConnect~\citep{mobiny2021dropconnect}, Deep Ensembles~\citep{lakshminarayanan2017simple} and Flipout~\citep{wen2018flipout} each construct different approaches to sample parameters $\theta$ from an approximation of $p(\theta|D)$, as described in Appendix \ref{sec:uq_methods}.

Equation \ref{eq:aleatoric_epistemic} gives a predictions informed by aleatoric and epistemic uncertainty, but it does not quantify each uncertainty. For this we describe the Gaussian Logits and Information Theoretic disentangling below.

\paragraph{Gaussian Logits Disentangling}\label{sec:gauss-logit-background}

Gaussian Logits disentangling (shown in Figure \ref{fig:arch_disentanglement}) follows from how disentangling works in regression. In heteroscedastic regression the model predicts a Gaussian distribution for a sample, instead of a single point. The model has two \textit{heads}, one for the mean $\mu(x)$, and another for the variance $\sigma^2(x)$. The variance learns the heteroscedastic (aleatoric) uncertainty \citep{seitzer2022pitfalls}. When this is combined with a Bayesian Neural Network we sample different parameters $\theta$, resulting in $T$ samples of predictions for $\mu_t(x)$ and $\sigma^2_t(x)$. AU is then estimated by the mean of the output variances $\mathbb{E}[\sigma^2_t(x)]$. While EU is estimated by the variance of the output means $\mathrm{Var}[\mu_t(x)]$ \citep{kendall2017uncertainties}. 

When this is applied to classification, the logits are estimated as a Gaussian distribution \citep{collier2023massively}. The variance $\bm{\sigma}^2(x)$ can be determined by either the aleatoric ($\bm{\sigma}^2(x) = \mathbb{E}[\bm{\sigma}^2_t(x)]$) or epistemic ($\bm{\sigma}^2(x) = \mathrm{Var}[\bm{\mu}_t(x)]$) uncertainty. Softmax is then applied to samples drawn from a Gaussian distribution $\mathbf{z} \sim \mathcal{N}(\bm{\mu}(x); \bm{\sigma}^2(x))$ \citep{valdenegro2022deeper}
\begin{equation} \label{eq:sampling_softmax}
    p(y|x) = \frac{1}{N} \sum_N \mathrm{softmax}(\mathbf{z}). 
\end{equation}
This results in either AU informed probabilities, or EU informed probabilities. We take the mean over the sampled probabilities to get a probability vector of length $C$. On these probabilities the entropy
\begin{equation}\label{eq:entropy}
    \mathbb{H}[p(y|x)] = -\sum_{c \in C} p(y=c|x) \log p(y=c|x)
\end{equation} 
  gives us a single value for uncertainty determined by either AU or EU depending on the variance selected. %

\begin{figure}[tbp]
    \centering
    \resizebox{\columnwidth}{!}{
        \begin{tikzpicture}[
    neuron/.style={circle, draw=darkblue!80, fill=white, inner sep=0pt, minimum size=14pt, thick, copy shadow={shadow xshift=1pt, shadow yshift=-1pt, fill=black!20}},
    box/.style={draw=darkblue, fill=mainblue!10, rounded corners=4pt, minimum width=1.4cm, minimum height=0.7cm, thick, font=\bfseries, copy shadow={shadow xshift=1.5pt, shadow yshift=-1.5pt, fill=black!15}},
    arrow/.style={-{Stealth[scale=1.2]}, thick, draw=black!70},
    label_style/.style={font=\footnotesize\sffamily\bfseries, color=black!60}
]

    \begin{scope}[on background layer]
        \fill[lightgray, rounded corners=10pt] (1.0, -2.5) rectangle (6.5, 2.5);
        \node[label_style, anchor=north] at (3.75, 2.4) {Hidden Layers};
    \end{scope}

    \node[box, fill=white] (X) at (-1.5, 0) {$X$};
    \node[label_style, anchor=north] at (X.south) {Input};

    \foreach \i in {1,...,5} \node[neuron] (L1-\i) at (1.8, \i*0.8 - 2.4) {};
    \foreach \i in {1,...,4} \node[neuron] (L2-\i) at (3.1, \i*0.8 - 2.0) {};
    \foreach \i in {1,...,5} \node[neuron] (L3-\i) at (4.4, \i*0.8 - 2.4) {};
    \foreach \i in {1,...,4} \node[neuron] (L4-\i) at (5.7, \i*0.8 - 2.0) {};

    \foreach \i in {1,...,5} \draw[arrow, thin, opacity=0.4] (X.east) -- (L1-\i.west);
    \foreach \i in {1,...,5} \foreach \j in {1,...,4} \draw[black!30, thin] (L1-\i) -- (L2-\j);
    \foreach \i in {1,...,4} \foreach \j in {1,...,5} \draw[black!30, thin] (L2-\i) -- (L3-\j);
    \foreach \i in {1,...,5} \foreach \j in {1,...,4} \draw[black!30, thin] (L3-\i) -- (L4-\j);

    \node[box] (mu) at (8.5, 0.7) {$\mu$};
    \node[box] (sigma) at (8.5, -0.7) {$\sigma$};

    \foreach \i in {1,...,4} {
        \draw[arrow, thin, opacity=0.4] (L4-\i.east) -- (mu.west);
        \draw[arrow, thin, opacity=0.4] (L4-\i.east) -- (sigma.west);
    }

    \begin{scope}[shift={(11.2, 0)}]
        \draw[thin, black!40] (-1, -1.0) -- (1, -1.0); %
        \draw[ultra thick, mainblue] (-0.8, -1.0) .. controls (-0.4, -1.0) and (-0.3, 0.6) .. (0, 0.6) 
                                       .. controls (0.3, 0.6) and (0.4, -1.0) .. (0.8, -1.0);
        \node[label_style, anchor=north, yshift=-3pt] (SampLabel) at (0, -1.0) {Sampling};
        
        \draw[arrow, dashed] (mu.east) -- (-0.8, 0.3);
        \draw[arrow, dashed] (sigma.east) -- (-0.8, -0.3);
    \end{scope}

    \draw[arrow] (12.2, 0) -- (13.5, 0);

    \begin{scope}[shift={(13.8, 0)}]
        \draw[thin, black!40] (0,-1.0) -- (0, 0.8); %
        \draw[thin, black!40] (0,-1.0) -- (1.2, -1.0); %
        \draw[ultra thick, mainblue] (0.1, -0.9) .. controls (0.5, -0.9) and (0.7, 0.7) .. (1.1, 0.7);
        \node[label_style, anchor=north, yshift=-3pt] at (0.6, -1.0) {Softmax};
    \end{scope}

\end{tikzpicture}
    }
    \caption{Diagram of Gaussian Logits disentangling.}
    \label{fig:arch_disentanglement}
\end{figure}

\paragraph{Information Theoretic Disentangling}\label{sec:information-theoretic-background}

In the Information Theoretic (IT) approach the predicted probabilities $p(y=c|x, \theta)$ are considered to represent AU. In this case a standard Softmax output is considered, but the multiple samples of parameters $\theta \sim \Theta$ still result in multiple samples of predicted probabilities. In this case, the entropy of the mean probability $\mathbb{H}[\mathbb{E}_\Theta[p(y|x, \theta)]]$ is considered to represent the total uncertainty, whereas the Expected Entropy of each probability vector $\mathbb{E}_\Theta[\mathbb{H}[p(y|x, \theta)]]$ represents the AU. The difference between them is considered a measure of EU and is an approximation of the Mutual Information $\mathbb{I}(Y; \Theta)$, where $Y$ is the Random Variable from which the label $y$ is drawn \citep{mukhoti2023deep}. This is practically approximated by assuming that the total uncertainty is the sum of the epistemic and aleatoric uncertainty  such that
\begin{equation}\label{eq:it_disentanling}
    \underbrace{\mathbb{I}(Y; \Theta)}_{Epistemic} \approx \underbrace{\mathbb{H}[\mathbb{E}_\Theta[p(y|x, \theta)]]}_{Total} - \underbrace{\mathbb{E}_\Theta[\mathbb{H}[p(y|x, \theta)]]}_{Aleatoric}.
\end{equation}
\section{Theoretical Framework}\label{sec:theory}

\usepgfplotslibrary{groupplots} %
\usetikzlibrary{pgfplots.groupplots} %

\begin{figure*}[t]
\centering

\begin{tikzpicture}
    \begin{groupplot}[
        group style={
                group name=my plots,
                group size=2 by 1,
                x descriptions at=edge bottom,
                horizontal sep=2cm,
                vertical sep=0.3cm,
        },
        footnotesize,
        height=5cm,
        width=7cm,
        xmin=1,
        xmax=100,
        ymin=0,
        xlabel=Dataset size \%,
        ylabel=Uncertainty,
        ytick=\empty,
        xtick={0, 10, 20, 30, 40, 50, 60, 70, 80, 90, 100},
        domain=0:100
    ]

        \nextgroupplot[title={\footnotesize Experiment 1.}, xlabel={Dataset size $\%$}]
        
            \addplot[c0,no marks, thick] {-log10(x) + 2.2};
            \addplot[c1,no marks, thick] {-0.6*sin(x*500)/x + 0.3};

        \nextgroupplot[title={\footnotesize Experiment 2.}, xlabel={Labels shuffled $\%$}]
        
            \addplot[c0,no marks, thick] {x * 0.05 + 10};
            \addplot[c1,no marks, thick] {x * 0.8 + 30};

    \end{groupplot}

        \begin{groupplot}[
        group style={
                group name=my plots,
                group size=2 by 1,
                x descriptions at=edge bottom,
                y descriptions at=edge right,
                y descriptions at=all,
                horizontal sep=2cm,
                vertical sep=0.3cm,
        },
        footnotesize,
        height=5cm,
        width=7cm,
        xmin=1,
        xmax=100,
        ymin=0,
        ylabel=\textcolor{c2}{Accuracy},
        axis y line=right,
        ytick=\empty,
        domain=0:100
    ]
    
        \nextgroupplot
            \addplot[c2,no marks, thick] {log10(x)};

        \nextgroupplot
            \addplot[c2,no marks, thick] {x* -0.8 + 95};

    \end{groupplot}

\end{tikzpicture}

\caption{Expected behavior for our proposed experimental setup. Experiment 1: As dataset size increases, \textcolor{c0}{epistemic uncertainty $U_e$} decreases while \textcolor{c1}{aleatoric uncertainty $U_a$} remains stable on average. The change in \textcolor{c0}{$U_e$} is captured by the change in \textcolor{c2}{accuracy} when this is caused by dataset size. Experiment 2: With increasing label noise, \textcolor{c1}{$U_a$} rises while \textcolor{c0}{$U_e$} remains relatively stable, reflecting the model’s awareness of inherent data noise. Change in \textcolor{c1}{$U_a$} is captured by \textcolor{c2}{accuracy} when this is caused by label noise.  %
}
\label{fig:expected-behaviour}
\end{figure*}

To formalize the requirements for successful uncertainty disentanglement, we define the latent true uncertainties $U_a, U_e$ and their respective estimators $u_a, u_e$. We denote $\rho(\cdot, \cdot)$ as a correlation measure and $x \approxprop y$ to indicate a substantial positive correlation. We propose that disentanglement is achieved if and only if the following four conditions are satisfied:
\begin{align*}
    &\text{Consistency}   & u_a &\approxprop U_a \label{eq:cond1} \tag{C1} \\
    &                     & u_e &\approxprop U_e \label{eq:cond2} \tag{C2} \\[0pt]
    &\text{Orthogonality} & u_a &\; \cancel{\approxprop}\;  U_e \mid U_a \; \cancel{\approxprop}\;  U_e \label{eq:cond3} \tag{O1} \\
    &                     & u_e &\; \cancel{\approxprop}\;  U_a \mid U_a \; \cancel{\approxprop}\;  U_e \label{eq:cond4} \tag{O2}
\end{align*}
While standard evaluation protocols focus almost exclusively on Consistency (e.g \citet{lahlou2021deup, hofman2024quantifying}), we identify a fundamental flaw in this restricted view, which we formalize below.

\begin{theorem}[The Total Uncertainty Trap]
Evaluating only \textbf{Consistency} (\ref{eq:cond1}, \ref{eq:cond2}) is insufficient for disentanglement, as these conditions can be satisfied by a non-disentangled estimator.
\end{theorem}
\textit{Intuition.} Consider the total predictive uncertainty $u_p = u_a + u_e$. Because $u_p$ is a composite of both sources, it will satisfy both $u_p \approxprop U_a$ and $u_p \approxprop U_e$ without performing any internal separation. This demonstrates that Consistency alone is not a sufficient condition for disentanglement; Orthogonality (\ref{eq:cond3}, \ref{eq:cond4}) is the necessary requirement for true estimator independence.

Furthermore, we address the misconception regarding correlated estimators, which was used to suggest disentanglement failure in previous literature. \citet{mucsanyi2024benchmarking} showed that for CIFAR10 $\rho(u_a, u_e)\geq0.88$, and for ImageNet-1k $\rho(u_a,u_e) \geq 0.78$, suggesting that disentanglement failed.

\begin{theorem}[Necessity of Correlation]
If the latent uncertainties $U_a$ and $U_e$ are correlated within a dataset, then a correlation between the estimators ($u_a \approxprop u_e$) is a necessary consequence of estimator validity.
\end{theorem}
This theorem implies that the metric $\rho(u_e, u_a)$ is an invalid proxy for disentanglement quality. On complex datasets like ImageNet-1k, where aleatoric noise and epistemic difficulty often co-occur, a high correlation between $u_e$ and $u_a$ is a sign of a well-calibrated model rather than a failure of separation. We conclude with a unified definition:

\begin{theorem}[Fundamental Disentanglement]
An estimator $u_a$ is an orthogonally disentangled estimator of $U_a$ if and only if conditions \eqref{eq:cond1} and \eqref{eq:cond3} are satisfied. By symmetry, the same holds for $u_e$ via \eqref{eq:cond2} and \eqref{eq:cond4}.
\end{theorem}
Proofs for the necessity and sufficiency of these criteria are provided in Appendix \ref{sec:proofs}.

\section{Experimental Setup}
\label{sec:experimental_setup}

In Appendix \ref{sec:definining_au_eu} we define true latent aleatoric and epistemic uncertainty. However, these uncertainties cannot be directly observed. We circumvent this problem by manipulating them independently through different experiments and observing their relative effects. By manipulating conditions we can inject additional $U_a$ or $U_e$ and therefore still measure correlations. We define a protocol to observe Consistency and Orthogonality based on easy to reproduce experiments and argue for each of these experiments what the expected behaviour of $U_a$ and $U_e$ is based on agreed upon definitions. To quantify this as a metric, we introduce the \textit{Uncertainty Disentanglement Error} (UDE) to easily compare different disentanglement methods, and models. \citet{wimmer2023quantifying} previously explored orthogonal disentanglement, primarily axiomatically, but also with experiments that aim to manipulate $U_e$ and $U_a$. We follow the logic from \citet{wimmer2023quantifying} that a decrease in accuracy under experiments that vary the dataset (epistemic) or add noise (aleatoric) indicates an increase in  $U_e$ or $U_a$ respectively. We expand upon their work by being selective in how ``noise" is added, as some of their experiments with noise may also introduce epistemic uncertainty (as we explain in Appendix \ref{sec:not-included-experiments}). An intuitive overview of these experiments alongside their rationale is presented in Figure \ref{fig:expected-behaviour}. This shows two experiments manipulating the size of the dataset, and the label noise in the dataset, effectively manipulating $U_e$ and $U_a$ respectively.

The protocol is applied to both disentangling methods, on multiple data domains and Bayesian Neural Network (BNN) approximations. We evaluate our criteria on CIFAR-10 \citep{hendrycks2019benchmarking} and Fashion MNIST \citep{xiao2017fashion} for computer vision, the UCI Wine dataset \citep{misc_wine_109} for tabular data, and a Brain-Computer Interface (BCI) dataset \citep{brunner2008bci} as a timeseries application where data quality is poor. On these datasets, we apply four different BNN approximations to observe method-dependent effects. Specifically, we compare MC-Dropout \citep{gal2016dropout} with $t=50$ forward passes, Deep Ensembles \citep{lakshminarayanan2017simple} with $t=10$ models, MC-DropConnect \citep{mobiny2021dropconnect} using a dropout probability of $p=0.3$ and $t=50$ forward passes, and Flipout \citep{wen2018flipout} substituting the fully connected layer with a Flipout layer using a prior $p(\theta) = \mathcal{N}(0, 5^2) + \pi ~ \mathcal{N}(0, 2^2)$ where $\pi = 0.5$. Flipout is trained for 500 epochs, while other models are trained for 100 epochs. Experiments are repeated five times for robustness, with implementations available in our repository\footnote{\url{https://anonymous.4open.science/r/uq_disentanglement_comparison-72CC}}, and detailed in Appendix \ref{sec:model_implementation}. Further empirical analyses exploring finetuning, dropout rate, and ImageNet-1k are described in Section \ref{sec:UDE}, and detailed in Appendices \ref{sec:dropout} and \ref{sec:imagenet_robustness}.

\begin{figure*}
    \begin{minipage}{0.48\linewidth}
             \begin{experimentbox}[label=exp:dataset]{Dataset Scaling}
    \textbf{Objective:} Vary the training dataset size $N$ to control accuracy and epistemic uncertainty.
    \vspace{2mm} 
    \\
    \textbf{Expected Outcome:} 
    \begin{itemize}[noitemsep, topsep=2pt, leftmargin=15pt]
        \item \textbf{Epistemic uncertainty ($U_e$)} should decrease monotonically when accuracy increases, as the parameter space gets more constrained.
        \item \textbf{Aleatoric uncertainty ($U_a$)} should remain a constant floor, as it represents the inherent noise.
    \end{itemize}
    \tcbline
    \footnotesize \textbf{Criteria:}  (\ref{eq:cond2}): $u_e \approxprop U_e$, (\ref{eq:cond3}): $u_a \; \cancel{\approxprop}\; U_e \mid U_a\; \cancel{\approxprop}\; U_e$
    \end{experimentbox}
    \end{minipage}\hfill
    \begin{minipage}{0.48\linewidth}
        \begin{experimentbox}[label=exp:noise]{Label Noise Injection}
    \textbf{Objective:} Increase aleatoric uncertainty $U_a$ by adding stochastic noise to training labels.
    \vspace{2mm} \\
    \textbf{Expected Outcome:} 
    \begin{itemize}[noitemsep, topsep=2pt, leftmargin=15pt]
        \item \textbf{Aleatoric uncertainty ($U_a$)} should scale proportionally with noise levels, as it is expected to learn the noise as aleatoric uncertainty.
        \item \textbf{Epistemic uncertainty ($U_e$)} should remain consistent, as it should not be affected by the noise levels except for extreme cases.
    \end{itemize}
    \tcbline
    \footnotesize \textbf{Criteria:} (\ref{eq:cond1}): $u_a \approxprop U_a$, (\ref{eq:cond4}): $u_e\; \cancel{\approxprop}\; U_a \mid U_a\; \cancel{\approxprop}\; U_e$
    \end{experimentbox}
    \end{minipage}
\end{figure*}

\subsection{Epistemic Uncertainty Experiment} 
\label{sec:dataset_size_experiment}

\begin{figure*}[b]
    \begin{minipage}{0.48\textwidth}
        \input{tikzfigures/dataset_size.tex}

    \end{minipage}\hfill
    \begin{minipage}{0.48\textwidth}
            \input{tikzfigures/label_noise.tex}
    \end{minipage}
\end{figure*}

Experiment \ref{exp:dataset} builds on the notion that $U_e$ can be reduced while $U_a$ cannot \citep{abdar2021review}. By manipulating the training data volume, we directly influence $U_e$ while $U_a$ stays constant, allowing us to observe the consistency of $u_e$ and the orthogonality of $u_a$. We re-train the models seven times using $1\%$, $5\%$, $10\%$, $25\%$, $50\%$, $75\%$, and $100\%$ of the data. Sub-sampling is done within each class to maintain balances \cite{valdenegro2021exploring}, and training epochs are inversely proportional to the data volume to prevent underfitting. In Appendix \ref{sec:underfitting-two-moons} we show results on the Two Moons dataset, showing that the model underfits if the number of epochs is not scaled inversely proportional. 

As shown in Figure \ref{fig:decreasing_dataset_CIFAR10}, aleatoric estimates $u_a$ consistently increase with larger datasets across all models, indicating that orthogonality criterion (\ref{eq:cond3}) is not met. Furthermore, $u_e$ does not always decrease as expected; under Gaussian Logits (GL) disentanglement, MC-Dropout and MC-DropConnect fail to show a consistent decrease, whereas Information Theoretic (IT) disentangling successfully meets consistency criterion (\ref{eq:cond2}). Deep Ensembles show the best accuracy and $u_e$ consistency ($\rho(u_e, U_e) > 0.95$). 

Feature-space visualizations on the Two Moons dataset (Figure \ref{fig:two-moons-dataset-size}) shows that IT and GL give qualitatively different behaviors as the dataset size increases.

\subsection{Aleatoric Uncertainty Experiment}
\label{sec:label_noise_experiment}

In Experiment \ref{exp:noise} we introduce additional $U_a$ by randomly swapping labels between $0\%$ and $100\%$ of the samples by steps of $10\%$. This ensures input features remain identical, theoretically keeping $U_e$ constant while manipulation of $U_a$, allowing us to test consistency of $u_a$ (\ref{eq:cond1}) and orthogonality of $u_e$ (\ref{eq:cond4}). 

\begin{figure*}[t]
\centering
\begin{subfigure}[t]{.45\textwidth}
  \centering
  \includegraphics[width=\linewidth]{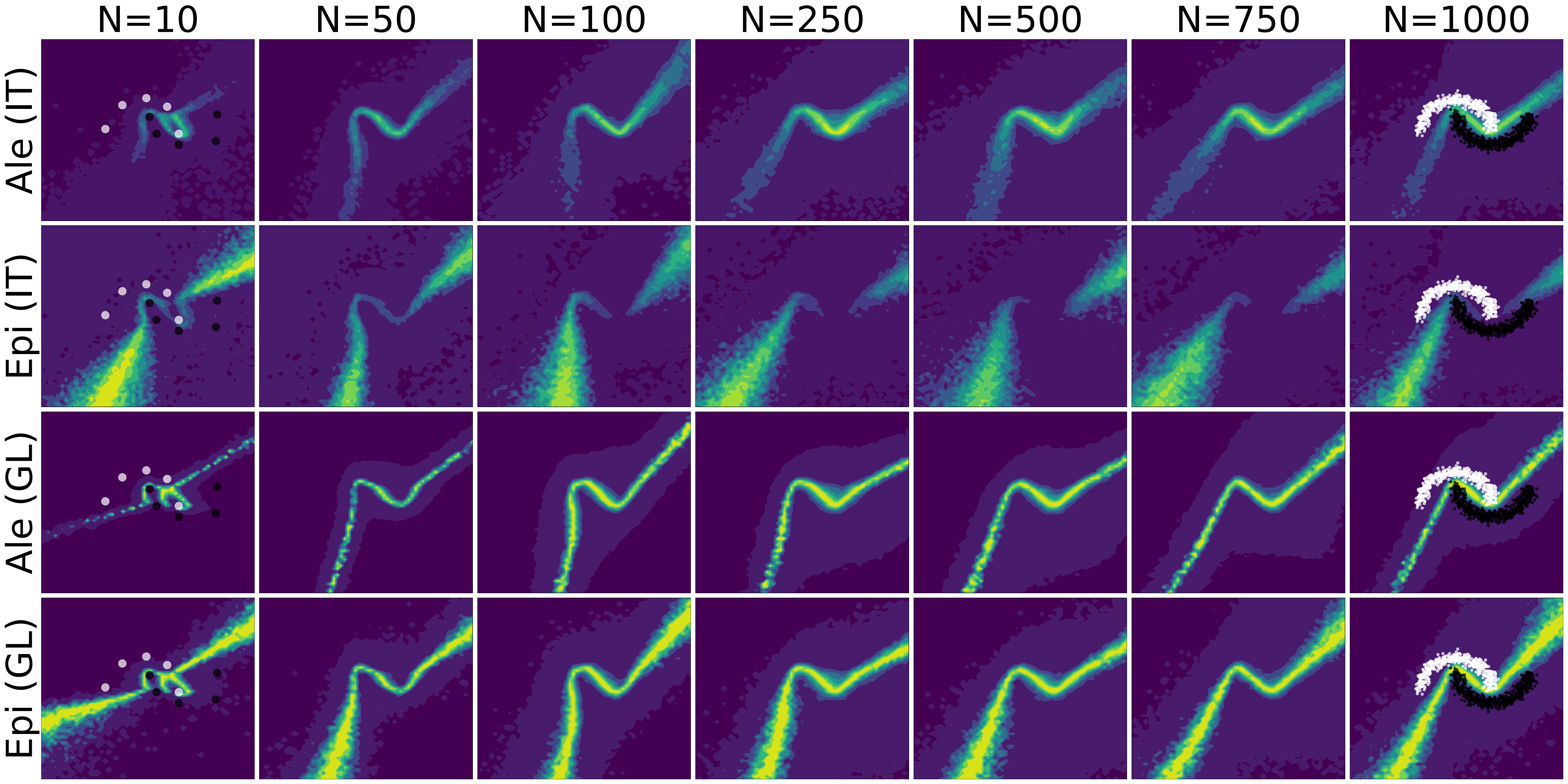}
  \caption{Decreasing Dataset}
  \label{fig:two-moons-dataset-size}
\end{subfigure}\hfill
\begin{subfigure}[t]{.45\textwidth}
  \centering
  \includegraphics[width=\linewidth]{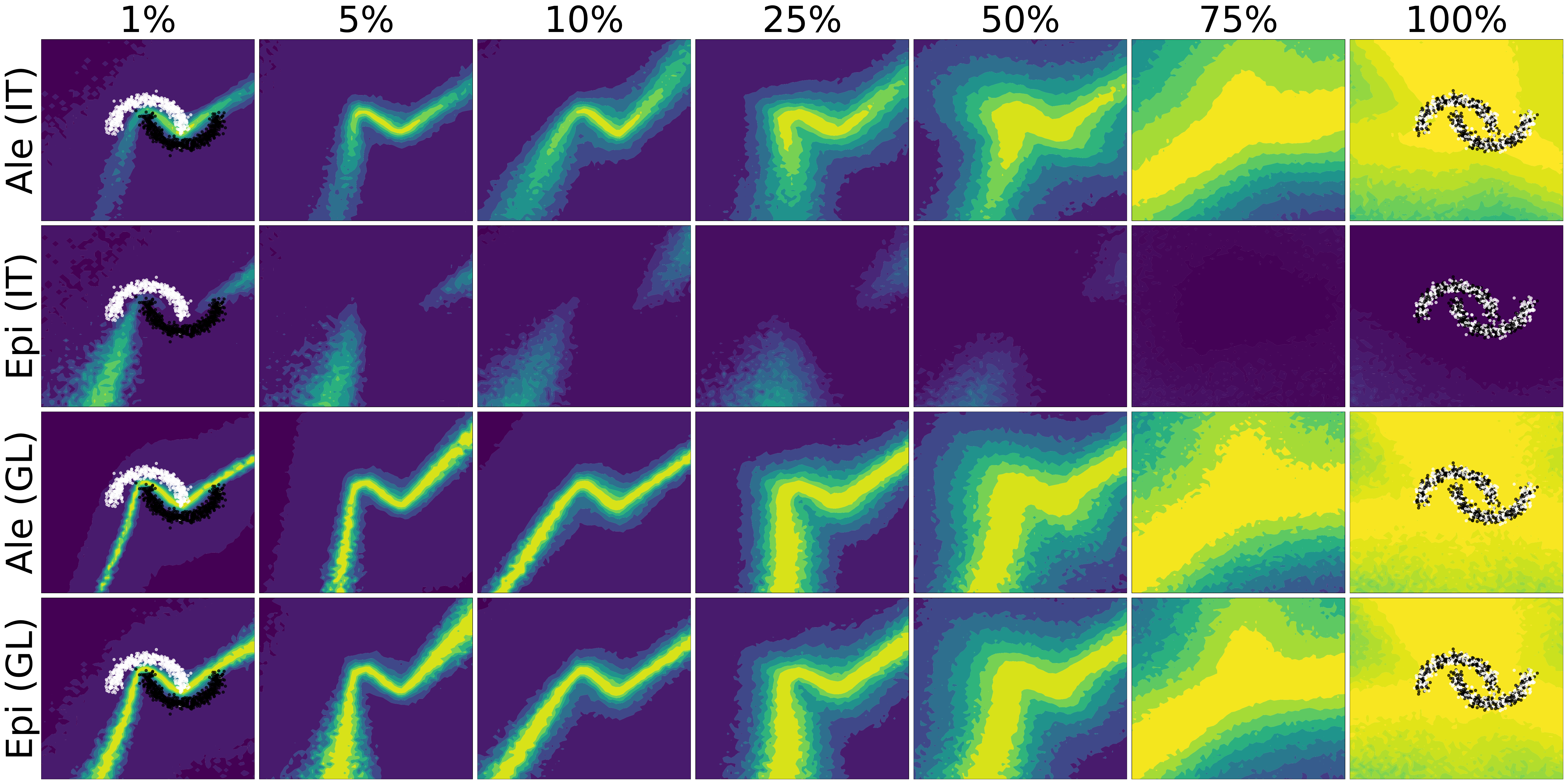}
  \caption{Label Noise}
  \label{fig:two-moons-label-noise}
\end{subfigure}\hfill
\caption{Aleatoric $u_a$ and epistemic uncertainty $u_e$ with (a) changing dataset sizes or (b) changing label noise for the Two Moons dataset with MC-Dropout. The lighter areas represent higher uncertainty. By visualizing the uncertainty for the whole feature space, we can gain intuition about uncertainty outside the dataset. Gaussian Logits gives qualitatively different results than Information Theoretic.}
\label{fig:two-moons}
\end{figure*}

Figure \ref{fig:label_noise_CIFAR10} demonstrates that as $U_a$ increases, the GL approach incorrectly increases both $u_a$ and $u_e$, failing orthogonality. In contrast, the IT approach keeps $u_e$ relatively consistent, but a minor correlation remains. At $100\%$ noise, IT shows decrease in $u_e$ as the learning task becomes meaningless and the decision boundary collapses (see Figure \ref{fig:two-moons-label-noise}).

\begin{table*}[b]
\centering
\renewcommand{\arraystretch}{1.2} 
\caption{UDE for different methods, models and datasets. \textbf{IT Disentangling} consistently achieves lower error and higher stability than \textbf{GL}. Green and yellow denote best and second-best results per dataset.}
\label{tab:DisentanglementScore}
\begin{tabular}{@{} l cccc >{\columncolor{sthlmGrey!15}}c @{}}
\toprule
\rowcolor{sthlmBlue!10} 
\textbf{Uncertainty Method} & \textbf{CIFAR10} & \textbf{F-MNIST} & \textbf{Wine} & \textbf{BCI} & \textbf{Average} \\ \midrule
\rowcolor{sthlmBlue!5}
\multicolumn{6}{l}{\textbf{GL Disentangling}} \\
MC-Dropout      & 0.661 $\pm$ .044 & 0.673 $\pm$ .034 & 0.778 $\pm$ .031 & 0.803 $\pm$ .040 & 0.729 $\pm$ .037 \\
MC-DropConnect  & 0.481 $\pm$ .006 & 0.399 $\pm$ .022 & 0.737 $\pm$ .080 & 0.638 $\pm$ .055 & 0.564 $\pm$ .041 \\
Flipout         & 0.418 $\pm$ .014 & 0.480 $\pm$ .004 & 0.490 $\pm$ .094 & 0.639 $\pm$ .054 & 0.507 $\pm$ .042 \\
Deep Ensembles  & 0.659 $\pm$ .055 & 0.439 $\pm$ .013 & 0.524 $\pm$ .054 & 0.737 $\pm$ .095 & 0.590 $\pm$ .054 \\ \midrule
\rowcolor{sthlmBlue!5}
\multicolumn{6}{l}{\textbf{IT Disentangling}} \\
MC-Dropout      & \cellcolor{green!20}0.295 $\pm$ .014 & \cellcolor{green!20}0.294 $\pm$ .032 & 0.645 $\pm$ .041 & 0.608 $\pm$ .062 & 0.460 $\pm$ .037 \\
MC-DropConnect  & 0.331 $\pm$ .033 & \cellcolor{yellow!20}0.320 $\pm$ .035 & \cellcolor{yellow!20}0.373 $\pm$ .070 & 0.811 $\pm$ .066 & 0.459 $\pm$ .051 \\
Flipout         & \cellcolor{yellow!20}0.312 $\pm$ .006 & 0.354 $\pm$ .006 & \cellcolor{green!20}0.337 $\pm$ .045 & \cellcolor{green!20}0.448 $\pm$ .046 & \cellcolor{green!20}0.363 $\pm$ .034 \\
Deep Ensembles  & 0.415 $\pm$ .025 & 0.334 $\pm$ .015 & 0.454 $\pm$ .021 & \cellcolor{yellow!20}0.531 $\pm$ .054 & \cellcolor{yellow!20}0.434 $\pm$ .034 \\
\bottomrule
\end{tabular}
\end{table*}

\subsection{Concluding Discussion} 
The aggregated results from both experiments on all datasets (detailed in Appendix \ref{sec:detailed-results}, summarized in Table \ref{tab:DisentanglementScore} and Figure \ref{fig:error_contributions}), indicate that while consistency is achievable with a proper choice of BNN and disentanglement method, none of the evaluated methods achieve perfect orthogonality. Gaussian Logits fails entirely on orthogonality ($\rho(u_e, U_a) > 0.725$), while Information Theoretic disentangling performs substantially better but still shows moderate leakage ($\rho(u_{a}, U_e) \approx 0.6$). Overall, IT disentanglement with Flipout or Deep Ensembles provides the most robust response to changes in true epistemic uncertainty, making it the preferable choice for disentangled uncertainty estimation.

\section{Uncertainty Disentanglement Error}\label{sec:UDE}
The experiments presented above evaluate whether aleatoric and epistemic uncertainty are consistently and orthogonally estimated. We show that different methods of disentanglement give qualitatively different behaviour. To combine these criteria into a single metric measuring orthogonal disentanglement, we introduce the Uncertainty Disentanglement Error (UDE). UDE captures the deviation between the observed and desired Pearson Correlation Coefficients $\rho(u_{(\cdot)}, U_{(\cdot)} )$ between $-acc$ and the uncertainty components. This metric is grounded in the necessary and sufficient criteria for disentanglement: in the Dataset Size experiment, epistemic uncertainty should be strongly correlated with error ($\rho(u_e, U_e) \rightarrow 1$), while aleatoric uncertainty should not correlate ($|\rho(u_a, U_e)| \rightarrow 0$). Conversely, in the Label Noise experiment, aleatoric uncertainty should correlate with error ($\rho(u_a, U_a) \rightarrow 1$), and epistemic uncertainty should not ($|\rho(u_e, U_a)| \rightarrow 0$). 

Based on this, the UDE is therefore defined as
\begin{align}   
\begin{split}
\text{UDE} =  \frac{1}{4} & (
\underbrace{|\rho(u_a, U_a)-1|)}_{\text{(\ref{eq:cond1})}}
 + \underbrace{|\rho(u_e, U_e)-1|)}_{\text{(\ref{eq:cond2})}}\\
  & + \underbrace{|\rho(u_a, U_e)|)}_{\text{(\ref{eq:cond3})}}
 + \underbrace{|\rho(u_e, U_a)|))}_{\text{(\ref{eq:cond4})}}.
\end{split}
\end{align}
A lower UDE score indicates better alignment with the conditions described in Section \ref{sec:theory} defining consistent and orthogonal uncertainty disentanglement. %

In Table \ref{tab:DisentanglementScore} we report the UDE for all models, disentanglement methods, and datasets presented, with detailed plots in Appendix \ref{sec:full_results}. This establishes a state-of-the-art of UDE, and shows that Information Theoretic disentangling is better than Gaussian Logits disentangling in all conditions. From the UDE we can also observe that Flipout typically gives relatively good disentanglement, and that disentanglement on the BCI dataset is particularly difficult, presumably due to high signal and label noise and small datasets.

\paragraph{Empirical analysis of UDE}

\begin{figure}[t]
    \centering
    \includegraphics[width=0.9\linewidth]{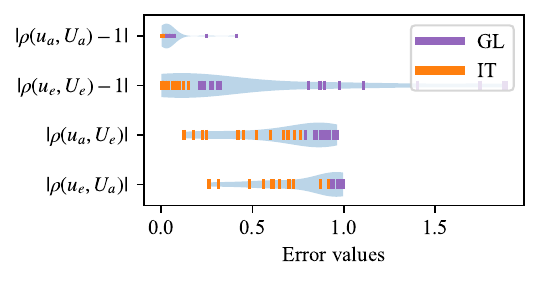}
    \caption{Values of each term in UDE, for the models and datasets shown in Table \ref{tab:DisentanglementScore}. Each tick indicates a run of a model on a dataset. This shows that both GL and IT disentangling have solved criterion (\ref{eq:cond1}), that IT has solved criterion  (\ref{eq:cond2}), but GL has not, and that neither method consistently comply with criteria  (\ref{eq:cond3}) and  (\ref{eq:cond4}).}
    \label{fig:error_contributions}
\end{figure}
To gain further insights in how UDE behaves and how it can be used in various contexts we perform further empirical analysis. 
Figure \ref{fig:error_contributions} shows the value of each term in UDE, corresponding to the results shown in Table \ref{tab:DisentanglementScore}. This shows that $|\rho(u_a,U_a)-1|$, corresponding to criterion (\ref{eq:cond1}), is effectively solved often achieving near zero error. We also see that $|\rho(u_e,U_e)-1|$, corresponding to criterion (\ref{eq:cond2}), also achieves low error for IT disentangling, but not for GL disentangling. The terms $|\rho(u_a,U_e)|$ and $|\rho(u_e,U_a)|$ quantifying orthogonality and corresponding to criteria (\ref{eq:cond3}) and (\ref{eq:cond4}) show substantial error. This shows that the consistency criteria are met with current methods, but orthogonality is not. This shows that there is a critical blindspot when disentanglement is evaluated only on consistency. 

In Table \ref{tab:ImageNetResults_maintext} we demonstrate UDE applied on ImageNet-1k, where we evaluated various alternatives to IT disentangling. To determine UDE for large-scale models, a more computationally affordable configuration is needed. We demonstrate the robustness of UDE, and how to adapt it to large models in Appendix 
\ref{sec:imagenet_robustness}. The results show that Information Theoretic disentangling is better than the ad-hoc Bias-Variance (B-V) disentangling \cite{smith2018understanding}, or disentangling with Pairwise KL-Divergence (KL) \cite{schweighofer2023introducing}. It also shows that for $u_e$, the IT disentangling effectively achieves both consistency and orthogonality, but $u_a$ only achieves orthogonality. In Appendix \ref{sec:dropout} we further explore how UDE can be optimized for finetuning the dropout-rate in MC-Dropout. This shows that UDE can be reliably optimized for, and that the choice of dropout rate can have a substantial impact on the consistency and orthogonality of $u_e$. %

\begin{table}[t]
\centering
\setlength{\tabcolsep}{3pt} %
\renewcommand{\arraystretch}{1.3} 
\caption{\textbf{ImageNet-1k results} from a Deep Ensemble trained from scratch. IT disentanglement achieves the lowest UDE.%
}
\label{tab:ImageNetResults_maintext}

\begin{tabular}{@{} l ccc @{}}
\toprule
\rowcolor{sthlmBlue!10} 
\textbf{Error Term} & \textbf{KL} & \textbf{B-V} & \cellcolor{sthlmBlue!20}\textbf{IT (Best)} \\ \midrule

\multicolumn{4}{l}{\textit{Target Consistency} ($\downarrow$)} \\
$|\rho(u_a, U_a)-1|$ & 0.013 & 0.013 & \cellcolor{green!20}0.001 \\
$|\rho(u_e, U_e)-1|$ & 0.027 & \cellcolor{green!20}0.025 & 0.026 \\ \midrule

\multicolumn{4}{l}{\textit{Cross-Leakage} ($\downarrow$)} \\
$|\rho(u_a, U_e)|$   & 0.953 & 0.953 & \cellcolor{green!20}0.947 \\
$|\rho(u_e, U_a)|$   & 0.687 & 0.635 & \cellcolor{green!20}0.015 \\ \midrule

\rowcolor{sthlmGrey!15}
\textbf{Total UDE}   & 0.419 & 0.407 & \cellcolor{green!20}0.248 \\
\bottomrule
\end{tabular}
\end{table}

\paragraph{Pretrained Models vs Models Trained from Scratch}
\begin{figure*}[tp]

    \begin{minipage}{.3\linewidth}
         \centering
            
            \pgfplotsset{scaled y ticks=false}
            
            \begin{tikzpicture}
            \begin{axis}[
                footnotesize,
                height=4.5cm,
                width=4.5cm,
                xlabel=Dataset size,
                ylabel=Uncertainty,
                xtick={0, 0.5, 1.0},
                yticklabel style={
                    /pgf/number format/fixed,
                    /pgf/number format/precision=5
                },
                ytick pos=left,
                ymin=0,
                ymax=1.3,
                ytick distance=0.2,
            ]
            
                \pgfplotstableread[col sep=comma]{./data/decreasing_dataset_results_de_it_finetuning_cifar10.csv}\rawdata
                
                \pgfplotstablesort[sort key={changed_parameter_values}]{\sorted}{\rawdata} %

                \addplot[name path=aleatoric, c1] 
                    table[x={changed_parameter_values}, y={aleatoric_uncertainties}]{\sorted};
                
                \addplot[name path=aleatoric_upper, draw=none]
                    table[x=changed_parameter_values,
                          y expr=\thisrow{aleatoric_uncertainties} + 2* \thisrow{aleatoric_uncertainties_std}]
                          {\sorted};
                
                \addplot[name path=aleatoric_lower, draw=none]
                    table[x=changed_parameter_values,
                          y expr=\thisrow{aleatoric_uncertainties} - 2*  \thisrow{aleatoric_uncertainties_std}]
                          {\sorted};
                
                \addplot[c1!30, fill opacity=0.3]
                    fill between[of=aleatoric_upper and aleatoric_lower];

                \addplot[name path=epistemic, c0]
                    table[x=changed_parameter_values, y=epistemic_uncertainties] {\sorted};
                
                \addplot[name path=epistemic_upper, draw=none]
                    table[x=changed_parameter_values,
                          y expr=\thisrow{epistemic_uncertainties} + 2* \thisrow{epistemic_uncertainties_std}]
                          {\sorted};
                
                \addplot[name path=epistemic_lower, draw=none]
                    table[x=changed_parameter_values,
                          y expr=\thisrow{epistemic_uncertainties} - 2* \thisrow{epistemic_uncertainties_std}]
                          {\sorted};
                
                \addplot[c0!30, fill opacity=0.3]
                    fill between[of=epistemic_upper and epistemic_lower];

            \end{axis}

            \pgfplotsset{scaled y ticks=false}

            \begin{axis}[
                footnotesize,
                height=4.5cm,
                width=4.5cm,
                ylabel={\color{c2}Accuracy},
                xtick=\empty,
                yticklabel style={
                    /pgf/number format/fixed,
                    /pgf/number format/precision=5
                },
                ymin=0.55,
                ymax=0.90,
                ytick distance=0.1,
                ytick pos=right
            ]
            
                \pgfplotstableread[col sep=comma]{./data/decreasing_dataset_results_de_it_finetuning_cifar10.csv}\rawdata
                \pgfplotstablesort[sort key={changed_parameter_values}]{\sorted}{\rawdata} %
                \addplot[name path=accuracy, c2]
                    table[x=changed_parameter_values, y=accuracies] {\sorted};

                \addplot[name path=accuracy_upper, draw=none]
                    table[x=changed_parameter_values,
                          y expr=\thisrow{accuracies} + 2*  \thisrow{accuracies_std}]
                          {\sorted};
                
                \addplot[name path=accuracy_lower, draw=none]
                    table[x=changed_parameter_values,
                          y expr=\thisrow{accuracies} - 2* \thisrow{accuracies_std}]
                          {\sorted};
            
                \addplot[c2!30, fill opacity=0.3]
                    fill between[of=accuracy_upper and accuracy_lower];
            
            \end{axis}
            \end{tikzpicture}

    \end{minipage}\hfill
    \begin{minipage}{.3\linewidth}
        
      \centering   
            \pgfplotsset{scaled y ticks=false}
            \begin{tikzpicture}
            \begin{axis}[
                footnotesize,
                height=4.5cm,
                width=4.5cm,
                xlabel=Label noise,
                ylabel=Uncertainty,
                xtick={0, 0.5, 1.0},
                yticklabel style={
                    /pgf/number format/fixed,
                    /pgf/number format/precision=5
                },
                ytick pos=left,
                ymin=0,
                ymax=2.4,
                ytick distance=0.5,
            ]
            
                \pgfplotstableread[col sep=comma]{./data/label_noise_results_de_it_finetuning_cifar10.csv}\rawdata
                
                \pgfplotstablesort[sort key={changed_parameter_values}]{\sorted}{\rawdata} %

                \addplot[name path=aleatoric, c1] 
                    table[x={changed_parameter_values}, y={aleatoric_uncertainties}]{\sorted};
                
                \addplot[name path=aleatoric_upper, draw=none]
                    table[x=changed_parameter_values,
                          y expr=\thisrow{aleatoric_uncertainties} + 2* \thisrow{aleatoric_uncertainties_std}]
                          {\sorted};
                
                \addplot[name path=aleatoric_lower, draw=none]
                    table[x=changed_parameter_values,
                          y expr=\thisrow{aleatoric_uncertainties} - 2*  \thisrow{aleatoric_uncertainties_std}]
                          {\sorted};
                
                \addplot[c1!30, fill opacity=0.3]
                    fill between[of=aleatoric_upper and aleatoric_lower];

                \addplot[name path=epistemic, c0]
                    table[x=changed_parameter_values, y=epistemic_uncertainties] {\sorted};
                
                \addplot[name path=epistemic_upper, draw=none]
                    table[x=changed_parameter_values,
                          y expr=\thisrow{epistemic_uncertainties} + 2* \thisrow{epistemic_uncertainties_std}]
                          {\sorted};
                
                \addplot[name path=epistemic_lower, draw=none]
                    table[x=changed_parameter_values,
                          y expr=\thisrow{epistemic_uncertainties} - 2* \thisrow{epistemic_uncertainties_std}]
                          {\sorted};
                
                \addplot[c0!30, fill opacity=0.3]
                    fill between[of=epistemic_upper and epistemic_lower];

            \end{axis}

            \pgfplotsset{scaled y ticks=false}            
            \begin{axis}[
                footnotesize,
                height=4.5cm,
                width=4.5cm,
                ylabel={\color{c2}Accuracy},
                xtick=\empty,
                yticklabel style={
                    /pgf/number format/fixed,
                    /pgf/number format/precision=5
                },
                ymin=0.10,
                ymax=0.90,
                ytick distance=0.1,
                ytick pos=right
            ]
            
                \pgfplotstableread[col sep=comma]{./data/label_noise_results_de_it_finetuning_cifar10.csv}\rawdata
                \pgfplotstablesort[sort key={changed_parameter_values}]{\sorted}{\rawdata} %
                \addplot[name path=accuracy, c2]
                    table[x=changed_parameter_values, y=accuracies] {\sorted};

                \addplot[name path=accuracy_upper, draw=none]
                    table[x=changed_parameter_values,
                          y expr=\thisrow{accuracies} + 2*  \thisrow{accuracies_std}]
                          {\sorted};
                
                \addplot[name path=accuracy_lower, draw=none]
                    table[x=changed_parameter_values,
                          y expr=\thisrow{accuracies} - 2* \thisrow{accuracies_std}]
                          {\sorted};
            
                \addplot[c2!30, fill opacity=0.3]
                    fill between[of=accuracy_upper and accuracy_lower];
            
            \end{axis}
        \end{tikzpicture}

    \end{minipage}\hfill
    \begin{minipage}{0.4\linewidth}

    \centering

    \setlength{\tabcolsep}{3pt} %
    \renewcommand{\arraystretch}{1.3} 
    \begin{tabular}{@{} l cc @{}}
    \toprule
    \rowcolor{sthlmBlue!10} 
    \textbf{Error Term} & \textbf{Pretrained}& \textbf{Orig.}\\\midrule

    $|\rho(u_a, U_a)-1|$             &   \cellcolor{green!20}0.025       & 0.073                       \\
    $|\rho(u_e, U_e)-1|$             &   0.266       & \cellcolor{green!20}0.042                       \\\midrule

    $|\rho(u_e, U_a)|$               &   0.985  &     \cellcolor{green!20}0.312                   \\
    $|\rho(u_a, U_e)|$               &   0.904  &     \cellcolor{green!20}-0.902           \\\midrule
    \textbf{Total UDE} &  0.545    &  \cellcolor{green!20} 0.332          \\\bottomrule
    \end{tabular}
    
    \end{minipage}
    \captionof{figure}{\textcolor{c0}{Epistemic uncertainty}, \textcolor{c1}{aleatoric uncertainty} and \textcolor{c2}{accuracy} for a ResNet18 Deep Ensemble pretrained on ImageNet-1k and finetuned to CIFAR10. Shaded areas indicate two standard deviations. We see overall small estimates of epistemic uncertainty. The pretrained model shows worse disentanglement than the model trained from scratch.}     \label{fig:finetuning_results}
\end{figure*}

Several works \cite{schweighofer2023introducing, mucsanyi2024benchmarking} evaluate the quality of aleatoric and epistemic uncertainty on pretrained models instead of models trained from scratch. We hypothesize that this may have an impact on the quality of disentanglement, and therefore, would pose a confound. %
We therefore compare our results on CIFAR10 in Section \ref{sec:experimental_setup} with results from a Deep Ensemble Resnet18 pretrained on Imagenet-1k. Each of the 3 models in the Deep Ensemble is finetuned from the same pretrained weights for 50 epochs. We used Information Theoretic disentangling because this is shown to achieve the best disentanglement in Table \ref{tab:DisentanglementScore}.

Figure \ref{fig:finetuning_results} shows the results of the orthogonality experiments. Overall, we see minimal estimates of epistemic uncertainty, much smaller than on models trained from scratch (Figures \ref{fig:decreasing_dataset_CIFAR10} and \ref{fig:label_noise_CIFAR10}). We also see $u_a$ decrease with increased accuracy $\rho(u_a, U_e) =0.985$ in the decreasing dataset experiment, which was negative in models trained from scratch (Table \ref{tab:dataset_size_correlations}, Appendix \ref{sec:detailed-results}).  The table shows the UDE and its components. UDE is substantially higher than in models trained from scratch, which is primarily due to the failed orthogonality, but partly due to the weaker consistency of $u_e$. Aleatoric and epistemic uncertainty seem to be almost fully conflated in this pretrained model. We conclude that there are substantial differences between models trained from scratch and pretrained models. This shows that findings on pretrained and from scratch models are not interchangeable. While \citet{mucsanyi2024benchmarking} found differences between CIFAR-10 and Imagenet-1k models, this may be because they used a pretrained model for ImageNet-1k, while their CIFAR-10 model was trained from scratch.

\section{Discussion}

Applications of uncertainty disentanglement methods assume that the predicted aleatoric and epistemic uncertainties can be used to pinpoint the origin of uncertainty \citep{vranken2021uncertainty, gill2021multicenter, van2022certainty, barandas2024evaluation}, but for this to work there should be no spurious interactions between aleatoric and epistemic uncertainty estimates ($u_a \; \cancel{\propto} \;U_e$ and $u_e \;\cancel{\propto}\; U_A$ if  $U_e \;\cancel{\propto}\; U_a$ ): they should be estimated orthogonally.

We have shown that only consistency is not sufficient, and that consistency and orthogonality together are sufficient and necessary conditions for disentanglement. Based on these requirements, we implemented experiments that manipulate the underlying ground-truth aleatoric and epistemic uncertainty, showing whether consistency and orthogonality requirements are met. These experiments are subsequently compiled into the Uncertainty Disentanglement Error metric, to quantify whether uncertainties are disentangled. We subsequently provided extensive empirical validation of these criteria. In doing so, we made the following main contributions:

\begin{itemize}[noitemsep, topsep=0pt]
    \item We prove that orthogonality is a necessary condition for disentanglement, that is not always met. This poses a blind-spot in standard methods for evaluating disentanglement. 
    \item We found that Deep Ensembles with Information Theoretic disentangling performs the best on consistency and orthogonality and is the current state-of-the-art for disentanglement. We show that on ImageNet-1k, $u_e$ is estimated orthogonally and consistently, but $u_a$ is not estimated orthogonally. This adds nuance to previous findings from \citet{mucsanyi2024benchmarking}, which suggested that disentanglement is insufficient because the estimators are correlated. We have shown that only $u_a$ fails orthogonality on ImageNet-1k, and that the estimates of $u_e$ in these conditions are consistent and orthogonal. 
    \item We have shown that disentanglement behavior is sensitive to whether models are trained from scratch or pretrained. We show that findings from pretrained models do not generalize to models trained from scratch.  
\end{itemize}

\paragraph{Limitations}

The primary limitation of this study is that it focuses on disentangled uncertainty at a dataset level, but minimally looks at individual samples. Good UDE on a dataset level is necessary for orthogonality at a sample level, but there is no guarantee it is sufficient. Future work manipulating the ground truth aleatoric and epistemic uncertainty for individual samples is needed to establish whether disentanglement works at the sample level. Such experiments may complement the currently presented dataset-level evaluation. 

Secondly, the proposed UDE assumes that epistemic and aleatoric uncertainty should each be linearly predictive of accuracy. This assumption is an extension of well-calibrated uncertainties \cite{guo2017calibration}, where the class probabilities (inverse of uncertainty) should perfectly match the probability of a correct prediction. From this follows that confidence (and therefore uncertainty) should have a linear relationship with accuracy. This assumption is also present in the formulation of aleatoric and epistemic uncertainty in Gaussian Logits disentangling \cite{valdenegro2022deeper}, and empirically holds both in our work and in \citet{mucsanyi2024benchmarking}. From this assumption of linearity also follows the additivity assumption, which is known to be problematic in classification \cite{wimmer2023quantifying}. Uncertainty measures that do not follow the additivity assumptions, and which admit a non-linear relationship to task performance may not be properly evaluated by UDE. For such models, variations of the UDE metric based on the same experiment (for example using rank-correlation) should be considered.

\section{Conclusion}
We find that neither Information Theoretic nor Gaussian Logits disentangling are able to separate aleatoric and epistemic uncertainty. Current methods for disentanglement have had the orthogonality criteria as a blind spot. We have show that orthogonality is necessary for disentanglement, and that this criteria is not always met. Using the Uncertainty Disentanglement Error we quantify consistency and orthogonality. Currently, the best method is Deep Ensembles trained from scratch, with Information Theoretic disentangling, showing consistent and orthogonal estimates of $u_e$ for ImageNet-1k, but the estimates for $u_a$ are not yet orthogonal.

\section*{Impact Statement}

Aleatoric and epistemic uncertainty have become well-known topics in uncertainty estimation, including methods to estimate them. However, outside of uncertainty-quantification experts, the limitations of such disentanglement are not as well presented in literature. Together with \citet{wimmer2023quantifying} and \citet{mucsanyi2024benchmarking} we want to point out that good uncertainty disentanglement is not a given. We propose UDE as a method to quantify whether estimates of aleatoric and epistemic uncertainty are really corresponding to the desired type of uncertainty. Such \textit{orthogonal} disentanglement is necessary to be able to make different choices under different types of uncertainty. 

Before implementing disentangled uncertainties (with low UDE) in a high-stakes situation or a decision support system as proposed by \cite{van2022certainty}, very close care needs to be taken to precisely define aleatoric and epistemic uncertainty in the context of the application. These definitions vary between experts \citep{kirchhof2025position}, and should be clearly communicated to users.

A realistic risk is that uncertainty disentanglement is commonly considered from a model perspective, but in larger pipelines this can break. For example a model might report more aleatoric uncertainty instead of epistemic uncertainty after strict regularization, but this nuance can easily be lost for a non-expert. Additionally, the users in a decision support system often have access to more features than the model, so humans may still make more informed estimates than a model with high aleatoric uncertainty and no epistemic uncertainty.

Disentangled uncertainties cannot be easily used in high-stakes scenarios and should be implemented with care and thoroughly evaluated in that specific scenario before being applied in practice. They are not ready to be trusted out-of-the-box.

\bibliography{reference.bib}
\bibliographystyle{icml2026}

\newpage

\onecolumn

\appendix

\section{Formalizing Orthogonal Disentanglement}
\label{sec:proofs}
We propose that good orthogonal disentanglement is achieved only when the following four conditions are satisfied±:
\begin{align*}
    &\text{Consistency}   & u_a &\approxprop U_a \tag{C1} \\
    &                     & u_e &\approxprop U_e \tag{C2} \\[5pt]
    &\text{Orthogonality} & u_a &\; \cancel{\approxprop}\;  U_e \mid U_a \; \cancel{\approxprop}\;  U_e \tag{O1} \\
    &                     & u_e &\; \cancel{\approxprop}\;  U_a \mid U_a \; \cancel{\approxprop}\;  U_e \tag{O2}
\end{align*}
Conditions \eqref{eq:cond1} and \eqref{eq:cond2} define \textit{consistency}: each estimate should reflect its corresponding true uncertainty. Conditions \eqref{eq:cond3} and \eqref{eq:cond4} define \textit{orthogonality}: each estimate should not reflect the alternative uncertainty source.

\subsection{Insufficiency of Consistency Metrics} \label{sec:proof_total_uncertainty}

\begin{theorem}[The Total Uncertainty Trap]
Evaluating only Conditions \eqref{eq:cond1} and \eqref{eq:cond2} does not evaluate disentanglement, as they can be satisfied by a non-disentangled estimator.
\end{theorem}

\begin{proof}
Consider a ``bad disentanglement'' scenario where the model fails to separate sources and instead outputs the total predictive uncertainty $u_p = U_a + U_e$ for both estimators, such that $u_a' = u_p$ and $u_e' = u_p$. While the identity $u_a' = u_e'$ already implies a lack of structural disentanglement, we demonstrate that such an estimator specifically fails the statistical orthogonality requirement.

To satisfy the consistency condition, we require a significant positive correlation $\rho(u_a', U_a) > 0$. We evaluate the covariance:
\begin{equation}
    \text{Cov}(u_a', U_a) = \text{Cov}(U_a + U_e, U_a) = \text{Var}(U_a) + \text{Cov}(U_e, U_a)
\end{equation}
Since $\text{Var}(U_a) > 0$ and assuming non-negative ground truth coupling ($\text{Cov}(U_e, U_a) \ge 0$), the numerator is strictly positive. Consequently:
\begin{equation}
    \rho(u_a', U_a) = \frac{\text{Var}(U_a) + \text{Cov}(U_a, U_e)}{\sqrt{\text{Var}(U_a + U_e)\text{Var}(U_a)}} > 0
\end{equation}
This confirms that $u_p$ ``passes" consistency without actual separation of sources. 

However, this same estimator fails the orthogonality requirement. To satisfy disentanglement, an estimate $u_a$ should not contain more information about $U_e$ than what is inherently shared between the ground truths. Following the assumption in Theorem A.3 that $\text{Cov}(U_a, U_e) > 0$, we evaluate the leakage:
\begin{equation}
    \text{Cov}(u_a', U_e) = \text{Cov}(U_a + U_e, U_e) = \text{Cov}(U_a, U_e) + \text{Var}(U_e)
\end{equation}
Because the estimator $u_a'$ incorporates the variance of the irrelevant source ($\text{Var}(U_e)$), it creates a spurious additive correlation such that:
\begin{equation}
    \rho(u_a', U_e) = \frac{\text{Cov}(U_a, U_e) + \text{Var}(U_e)}{\sqrt{\text{Var}(u_a')\text{Var}(U_e)}} > \rho(U_a, U_e)
\end{equation}

This confirms the estimator is entangled by leakage, as it fails to isolate the aleatoric signal even when accounting for the natural coupling between $U_a$ and $U_e$.
\end{proof}

\subsection{Necessity and Sufficiency}

\begin{theorem}[Necessity and Sufficiency]
An estimator $u_a$ is an orthogonally disentangled estimator of $U_a$ if and only if conditions \eqref{eq:cond1} and \eqref{eq:cond3} are satisfied.
\end{theorem}

\begin{proof}
\textbf{Necessity:} Suppose $u_a$ is an orthogonally disentangled estimator. By definition, $u_a$ must reflect the state of $U_a$, which implies $\rho(u_a, U_a) > 0$. Furthermore, orthogonality implies that $u_a$ is independent of $U_e$ except for correlation mediated by $U_a$. Thus, if $\text{Cov}(U_a, U_e) = 0$, then $\rho(u_a, U_e)$ must be $0$, satisfying Condition \eqref{eq:cond3}.

\textbf{Sufficiency:} Let $u_a = \alpha U_a + \beta U_e + \epsilon$. 
1. Condition \eqref{eq:cond1} ($\rho(u_a, U_a) > 0$) implies $\alpha \neq 0$.
2. Condition \eqref{eq:cond3} states that if $\text{Cov}(U_a, U_e) = 0$, then $\rho(u_a, U_e) = 0$. Evaluating:
\begin{equation}
    \text{Cov}(u_a, U_e) = \text{Cov}(\alpha U_a + \beta U_e + \epsilon, U_e) = \beta \text{Var}(U_e)
\end{equation}
For this to be zero while $\text{Var}(U_e) > 0$, we must have $\beta = 0$. Since $\alpha \neq 0$ and $\beta = 0$, $u_a$ is purely a function of $U_a$ and noise.
\end{proof}

\subsection{The Logical Fallacy of Estimator Correlation}

\begin{theorem}[Correlation is Necessary but Not Sufficient]
If the true uncertainties $U_a$ and $U_e$ correlate, then $u_a \approxprop u_e$ is a necessary consequence of estimator validity.
\end{theorem}

\begin{proof}
Assume ideal disentangled estimators $u_a = \alpha U_a + \epsilon_a$ and $u_e = \beta U_e + \epsilon_e$ ($\alpha, \beta > 0$). If $U_a \approxprop U_e$, then $\text{Cov}(u_a, u_e) = \alpha \beta \text{Cov}(U_a, U_e)$. Because $\alpha, \beta > 0$, the estimators must correlate. However, as shown in the proof of Theorem 1, the maximally entangled estimator $u_p$ also yields high correlation. Therefore, $u_a \approxprop u_e$ fails to distinguish between a perfect model and a fully entangled one.
\end{proof}

\section{Defining Aleatoric and Epistemic Uncertainty}\label{sec:definining_au_eu}
There are varying definitions for aleatoric and epistemic uncertainty \cite{kirchhof2025position}. This leads to confusion, for example about the concept of having a \textit{true} aleatoric or epistemic uncertainty. We define these concepts formally to agree on how these are considered within the scope of this paper, though definitions vary within the field of Uncertainty Quantification.

We define aleatoric uncertainty $U_a$ by considering a \textit{stochastic} function $f(x)$ which can be separated into a deterministic component $f'(x)$ and heteroscedastic noise $U_a(x)$ such that 
$$f(x) = f'(x) + U_a(x).$$ 
From this function we draw samples $((x_1, y_1), ..., (x_n, y_n))_{i=0, ..., n}$, allowing us to train a model using empirical risk minimization.
Then the epistemic uncertainty is the difference between the model and the true function.
$$U_e(x)= |f^*(x) - f'(x)|$$
$U_a$ is therefore inherent and irreducible, while $U_e$ can be reduced with more training samples or better priors. Since the true function $f(x)$ is not known, the aleatoric and epistemic uncertainty cannot be observed directly. Under this definition, out-of-distribution samples should still be samples from $f(x)$, but in parts of the domain that were not sampled in the training data, leading to more disagreement of the model, and therefore epistemic uncertainty.

\section{Background on Bayesian Neural Networks approximations}\label{sec:uq_methods}

For completeness we describe the workings of the BNN approximations used in this work. All these methods build on the assumptions that we can measure epistemic uncertainty by learning a distribution $\Theta$ from which to sample likely model parameters $\theta$, instead of learning a single optimal $\hat{\theta}$. To make predictions, all of these BNN approximations sample parameters $\theta \sim \Theta$, to produce a posterior distribution over predictions $f_\theta(x)$. 

\paragraph{Flipout} \cite{wen2018flipout} is the closest practical implementation of this. With Flipout each weight is represented by a mean and a Gaussian distributed perturbation. This is equivalent to sampling weights from a distribution $\theta \sim \mathcal{N}(\hat{\theta}, \hat{\sigma})$, where both the weights and the variances are learned through backpropagation using the reparameterization trick. When applied to deep models these variances stack up and can introduce an \textit{exploding variance} problem. To resolve this, Flipout is often (including in the current work) only applied to the last layers of the model. 

\paragraph{MC-DropConnect} \cite{mobiny2021dropconnect} similarly relies on perturbations to each weight. With MC-DropConnect weights are randomly set to 0. This is equivalent to multiplying each weight with a Bernoulli distribution. This operation is applied both during training and inference. Typically, the DropConnect layers are applied in the deeper layers of the model, though they can theoretically be applied anywhere without introducing instability. 

\paragraph{MC-Dropout} \cite{gal2016dropout} is very similar to DropConnect, but sets whole nodes to 0 instead of individual weights. This is based on the popular Dropout \cite{wager2013dropout} regularization method. MC-Dropout is a popular BNN approximation because it is easy to implement, does not negatively impact model accuracy, and can often be used in pre-trained models that were trained with Dropout regularization. Because MC-Dropout and MC-DropConnect are all applied during training, each of the sampled models can be considered likely for the given dataset. 

\paragraph{Deep Ensembles} \cite{lakshminarayanan2017simple} uses a computationally expensive method to generate few, but very good model samples. Deep Ensembles trains multiple instances of the same architecture on the same data, with only different random initialization. An ensemble of 5-10 models typically gives good performance. Deep Ensembles are commonly considered the state-of-the-art Bayesian Neural Network approximation for uncertainty quantification \cite{mucsanyi2024benchmarking}.

\section{Model Architectures and Pre-processing}\label{sec:model_implementation}

We implemented different model architectures following the same structure. All models ended in fully connected layers, where the Bayesian implementations would be applied. The Convolutional Neural Networks (for CIFAR10, Fashion MNIST and the BCI dataset) would have convolutional layers before this. For each dataset, we chose to use simple and established model architectures because the aim of this paper is not to achieve the highest performance, but to gain general insights into the behaviour of predicted aleatoric and epistemic uncertainty. All of the models are trained with the Adam optimizer with a learning rate of 1e-3, and a batch size of 128. 

The CIFAR10 and Fashion MNIST models followed the same setup. They use CNNs with 3 convolutional layers, each with a kernel size of $3 \times 3$, 64 filters and a \textit{relu} activation function. Each convolutional layer is followed by a $2 \times 2$ max pooling operation. This convolutional block is followed by a fully connected layer with 64 neurons with a dropout probability of $p=0.3$. This model performs reliably on both tasks. No preprocessing was needed on these datasets, and the original train-test split as provided was maintained throughout the experiment. 

For the Wine dataset we used a Multi-Layer Perceptron, with two non-Bayesian hidden layers of 32 nodes each, and one Bayesian hidden layer of 16 nodes. Since the Wine dataset has 13 features and 3 classes this gives and architecture of $13\times32\times32\times16\times3$. The Wine dataset underwent minimal preprocessing. 20\% of the dataset was used as test data, and the features were normalised using Z-score normalisation. 

The Motor Imagery BCI dataset \citep{brunner2008bci} is not a standard Machine Learning benchmark, and therefore requires specialised data handling (as provided by \citet{Aristimunha_Mother_of_all_2023}) and a specialised model architecture (based on \citep{manivannan2024uncertainty}). The dataset contains recordings from nine different subjects. For each subject a new model is trained and evaluated on that test subjects data. This gives us nine repetitions, instead of the five repetitions we used for other datasets. The 22 EEG channels are bandpass filtered between 7.5-30Hz, and downsampled to 128Hz. Each sample is a section of six seconds, during which a fixation cross is shown, followed by a pointing arrow. Based on the direction of the arrow, the subject will then perform one of four motor imaginations. 

The model architecture starts with a $1\times13$ temporal convolution with 40 kernels, followed by a $22\times1$ spatial convolution, again with 40 kernels, all with ReLU activation. After this, BatchNormalisation is applied, followed by a square activation function, 1x35 temporal average pooling with a stride of $1\times7$ and a log activation function. After this a fully connected Bayesian layer with 32 nodes connects to the output layer. This architecture gives performance in line with other models applied to this dataset \citep{manivannan2024uncertainty}. 

All experiments were performed on a dedicated model training server with two NVIDIA GeForce RTX 3090 GPUs, 64GB RAM, and a 12th Gen Intel Core i9 24-core CPU. Each experiments for each model (except Deep Ensembles) on all datasets takes approximately one day on this system. The total compute time for all results in this paper is therefore roughly 16 days.

\section{Alternative experiments} \label{sec:not-included-experiments}

Our paper focuses on three experiments that we consider robust in establishing the quality of disentanglement. There are many other ways in which the quality of aleatoric and epistemic uncertainty may be measured, but they all have limitations in terms of disentanglement. In most of these alternative experiments a manipulation to the ground-truth AU may also have a large effect on the ground-truth EU. Therefore, they cannot be reliably used for assessing the quality of disentanglement. We outline potential alternatives and the reason we do not use them below.

\paragraph{Datasets with known aleatoric uncertainty}
Some datasets where multiple people annotated a sample sometimes indicate a measure of annotator-disagreement. For example, the FER+ dataset \citep{BarsoumICMI2016} has emotion-annotated images from 10 different annotators. The disagreement between annotators establishes the inherent ambiguity in the classification task and is therefore a measure of ground-truth AU.

However, the ambiguity may also be substantially harder to learn. The disagreement between annotators may stem from ambiguity in emotions (aleatoric), or because it can be challenging (though not impossible) to find the identifying features of an emotion in an image (epistemic). Since this cannot be guaranteed at the ground-truth level, it also should not be used to assess the disentanglement.

\paragraph{Dataset shift}
Introducing EU by modifying the test data is a well established way to measure the quality of EU estimation \citep{ovadia2019can}. For measuring the quality of disentanglement however, this manipulation is not allowed to affect AU. Since these corruptions are usually made to be natural to the task (e.g. JPEG compression artifacts \citep{hendrycks2019benchmarking}) a model may have learned through AU that jpeg-artifacts indicate (aleatoric) uncertainty. 

While it may be possible to come up with corruptions that are difficult to learn from the \textit{clean} data, it is impossible to guarantee that they are not related. 

\paragraph{Added input noise in training data}
The Label Noise experiment adds noise to the training labels to introduce AU. It may be considered that noise can also be introduced at the input level by adding image corruptions \citep{hendrycks2019benchmarking, wimmer2023quantifying}. However, it is unclear whether this kind of corruptions really makes the relationship between the features and the labels more stochastic (aleatoric), or only more complex (epistemic). Therefore, it cannot be used to reliably evaluate disentanglement. 

\paragraph{Epistemic uncertainty should be better at Active Learning}
Since EU indicates that the uncertainty for a given sample can still be reduced, it is theoretically well suited for Active Learning. Normally, Active Learning considers the total uncertainty about a sample to identify whether it would benefit from learning its annotation. By considering only the epistemic aspect, the samples that are impossible to learn anyway (due to AU) are not selected. 

However, various studies have shown that aleatoric and epistemic uncertainty have a tendency to correlate \citep{valdenegro2022deeper, mucsanyi2024benchmarking}. Because of this, a good estimation of AU may be closer to the ground truth EU than a bad estimation of the EU. This is not a limitation of the disentangling, but only of the quality of EU estimation. 

This makes comparing an aleatoric-uncertainty based Active Learning strategy against an epistemic-uncertainty based Active Learning strategy an unreliable test for the quality of disentanglement.

\subsection{Out-of-Distribution Detection Experiment}

One of the alternative experiments to evaluate the quality of disentanglement relies on Out-of-Distribution detection. This is not included in the calculation of UDE, because we empirically find that aleatoric and epistemic uncertainty behave very similar on this task.  The motivation, methods, results and interpretation of this experiment are described below. 

The textbook example of uncertainty disentanglement has AU where there is noisy training data, and EU when moving away from the training data. For classification on toy data this is often shown using the Two Moons dataset, as we also saw in Figures \ref{fig:two-moons-dataset-size} and \ref{fig:two-moons-label-noise}. 

However, for high dimensional datasets, we often use OoD samples to observe high EU as they are away from the training data. For \textit{Soft-OoD} samples such as artificially corrupted images \citep{kotelevskii2024predictive} it cannot be guaranteed that the ground truth AU does not also increase, as a model can learn that blurry images have more AU. Instead, we focus on \textit{Hard-OoD} samples that come from a class that has not appeared in the training data. For these samples, AU predictions should be meaningless as the model will not have learned AU for these \citep{mukhoti2023deep}, but EU should be high as we are away from the training data. We should therefore expect that we can use EU to separate samples from an OoD class from the test samples of the ID class. AU should give arbitrary generalizations based on the AU in the training data, and therefore not be able to separate ID from OoD. 

\paragraph{Methods} The increase in uncertainty can be quantified through the ROC-AUC of separating the in-distribution (ID) classes from the OoD class by applying a threshold to the uncertainty \citep{barandas2024evaluation}. AU should not be able to separate the ID from OoD and maintain an ROC-AUC of around 0.5, while the ROC-AUC for EU should be higher.  
To put this in practice we remove one class from the training data and train the model only on the remaining classes. Then, we make predictions on the test data with all classes, where the different uncertainties are considered as a prediction for whether a sample is OoD. This approach is applied with each class left out once. %

\begin{table*}[t!]
\centering
\renewcommand{\arraystretch}{1.2} 
\caption{ROC-AUC for OoD class detection. Values are reported as \textbf{AU / EU}. EU should have a high ROC-AUC (closer to 1.0), while AU should have ROC-AUC near 0.5. Green and yellow denote best and second-best EU results per dataset.}
\label{tab:ood_roc_results_styled}
\begin{tabular}{@{} l cccc >{\columncolor{sthlmGrey!15}}c @{}}
\toprule
\rowcolor{sthlmBlue!10} 
\textbf{Uncertainty Method} & \textbf{CIFAR10} & \textbf{F-MNIST} & \textbf{Wine} & \textbf{BCI} & \textbf{Average} \\ \midrule
\rowcolor{sthlmBlue!5}
\multicolumn{6}{l}{\textbf{GL Disentangling}} \\
MC-Dropout      & 0.644 / 0.642 & 0.753 / 0.769 & 0.971 / 0.961 & 0.517 / 0.512 & 0.721 / 0.721 \\
MC-DropConnect  & 0.650 / 0.657 & 0.748 / 0.780 & 0.959 / 0.957 & 0.510 / 0.509 & 0.717 / 0.726 \\
Flipout         & 0.626 / 0.629 & 0.649 / 0.673 & 0.981 / 0.981 & 0.512 / 0.514 & 0.692 / 0.699 \\
Deep Ensembles  & \cellcolor{green!20}0.679 / 0.709 & \cellcolor{green!20}0.768 / 0.811 & \cellcolor{green!20}0.985 / 0.984 & \cellcolor{green!20}0.514 / 0.516 & \cellcolor{green!20}0.737 / 0.755 \\ \midrule
\rowcolor{sthlmBlue!5}
\multicolumn{6}{l}{\textbf{IT Disentangling}} \\
MC-Dropout      & 0.651 / 0.649 & 0.761 / 0.764 & 0.943 / 0.670 & 0.512 / 0.511 & 0.717 / 0.649 \\
MC-DropConnect  & 0.657 / 0.658 & \cellcolor{yellow!20}0.766 / 0.746 & 0.954 / 0.883 & \cellcolor{yellow!20}0.517 / 0.510 & 0.724 / 0.699 \\
Flipout         & 0.625 / 0.579 & 0.661 / 0.579 & \cellcolor{yellow!20}0.982 / 0.974 & 0.510 / 0.505 & 0.695 / 0.659 \\
Deep Ensembles  & \cellcolor{yellow!20}0.689 / 0.701 & 0.780 / 0.787 & 0.981 / 0.952 & \cellcolor{green!20}0.523 / 0.522 & \cellcolor{yellow!20}0.743 / 0.741 \\
\bottomrule
\end{tabular}
\end{table*}

\paragraph{Results} The ROC-AUC scores in Table \ref{tab:ood_roc_results_styled} show that both aleatoric and epistemic uncertainty increase for OoD samples, resulting in high ROC-AUC scores. This is contrary to the expectation that AU should not respond to the OoD samples. We find that EU estimates from the IT approach are actually worse than the AU estimates, suggesting that OoD detection actually benefits largely from aleatoric uncertainty. We also find that Flipout performs poorly, which is explained by its poor estimates of aleatoric uncertainty from the Label Noise experiment.

\paragraph{Conclusion} Since AU performs surprisingly well for both disentanglement approaches, and all uncertainty quantification methods, it is unlikely that this is a problem with the specific disentanglement. Instead, we consider that the learned mapping from a high dimensional input space to a lower dimensional hidden representation collapses the ID and OoD regions into the same hidden space, as suggested in \citep{shen2024uncertainty}. %
Based on these results, we consider that the OoD-detection experiment may be a good part of holistically evaluating the quality of uncertainty disentanglement, but is not suited to be used in computing the Uncertainty Disentanglement Error.

\FloatBarrier

\newpage

\section{Optimizing Dropout Probability for Uncertainty Disentanglement Error}\label{sec:dropout}

\begin{figure*}[t]
\centering
\begin{subfigure}[t]{.3\textwidth}
  \centering
  \includegraphics[height=5cm]{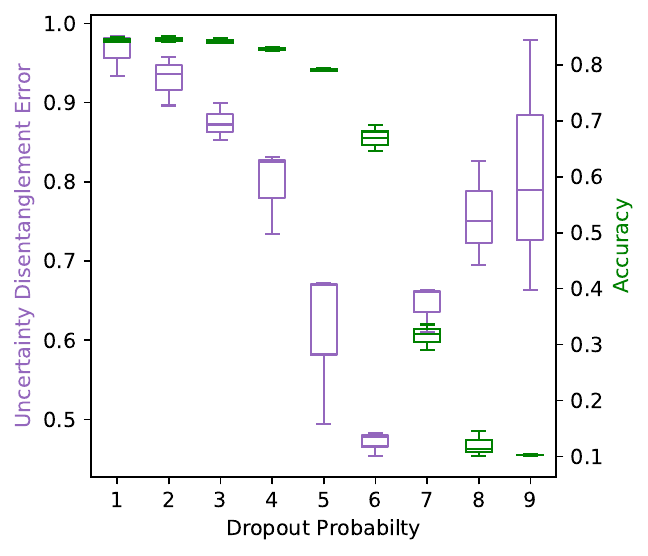}
  \caption{Uncertainty Disentanglement Error and accuracy}
  \label{fig:ude-dropout-p-total}
\end{subfigure}\hfill
\begin{subfigure}[t]{.6\textwidth}
  \centering
  \includegraphics[height=5cm]{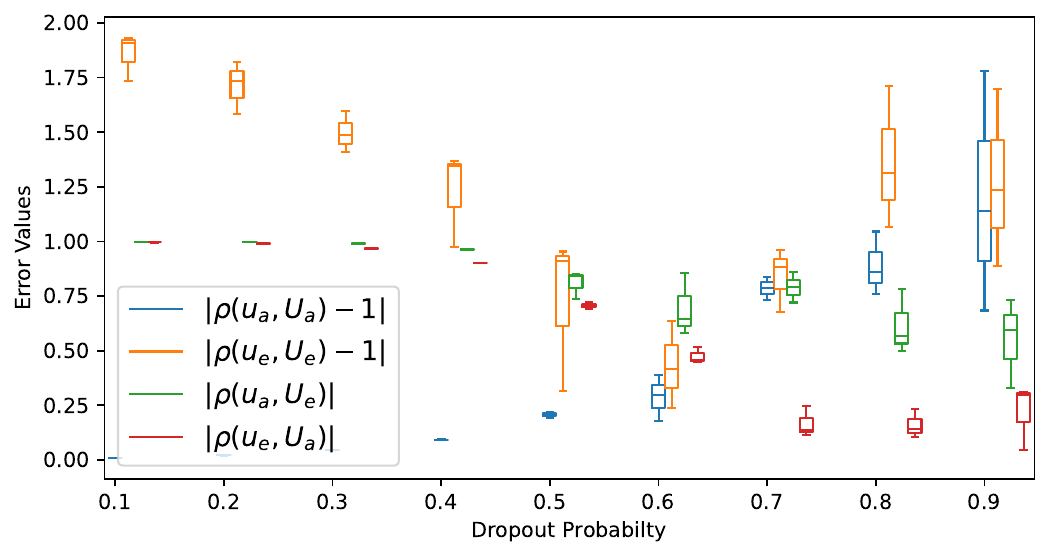}
  \caption{Individual Error Terms}
  \label{fig:ude-dropout-p-terms}
\end{subfigure}
\caption{UDE can be optimized by the choice of hyperparameters. (a) shows that UDE is minimal for $p=0.6$. Note that for higher dropout probability, the accuracy also decreases, causing a trade-off between UDE and accuracy. (b) shows how the individual UDE terms respond.. }\label{fig:dropout-p-ude}
\end{figure*}

We argue that UDE is something to optimize for. To demonstrate UDE as a reliable and clear metric to optimize for, we demonstrate a simple hyperparameter optimization case with CIFAR-10. 
We consider an MC-Dropout model with Information Theoretic disentangling, adding Dropout layers after every convolutional layer in a Resnet18 model. We optimize the dropout probability $p$ in the set $\{0.1, 0.2, ..., 0.9\}$. For each dropout probability, we calculate the UDE three times, allowing us to optimize the dropout probability for UDE specifically. Figure \ref{fig:dropout-p-ude} shows how UDE can be affected by the choice of the dropout probability. In this case, UDE is minimal with $p=0.6$. For $p\leq 0.5$, there is no substantial change in accuracy. From this analysis, $p=0.5$ can be selected, optimizing for accuracy and UDE. Of course this is not a general finding, and the optimal choice will be different for different models and datasets. More detailed insights can be gained from studying the individual components in Figure \ref{fig:ude-dropout-p-terms}.

\FloatBarrier

\section{Uncertainty Disentanglement Error for ImageNet-1k}\label{sec:imagenet_robustness}

Calculating the Uncertainty Disentanglement Error with the parameters described in Sections \ref{sec:dataset_size_experiment} and \ref{sec:label_noise_experiment} is computationally expensive. As described there, calculating the average Uncertainty Disentanglement Error requires re-training a model 54 times. With computationally expensive models like Deep Ensembles, this is equivalent to re-training a regular model 540 times. This is not feasible for large datasets and models that take one or more days to train. 

To compare uncertainty measures on ImageNet-1k \cite{russakovsky2015imagenet}, we first investigate how robust Uncertainty Disentanglement Error is with fewer training steps, using CIFAR10 as a heuristic. Specifically, we aim to compare variations of uncertainty measures based on Information Theoretic disentangling. We test the UDE of disentangling based on Equation \ref{eq:it_disentanling} from Section \ref{sec:background}, Pairwise-KL Divergence disentangling \cite{schweighofer2023introducing} where $u_e$ is estimated on the pairwise KL-divergence between models, and bias-variance disentangling \cite{smith2018understanding} where aleatoric uncertainty is based on the average probability, and epistemic uncertainty is based on variance between models. 

Figure \ref{fig:robustness_it_alternatives} shows how the UDE changes for these three pairs of uncertainty measures using a Deep Ensemble Resnet18 \cite{he2016deep} of 3 models trained on CIFAR10. This is repeated 3 times to get estimates for a standard deviation. From these results, we can see that the impact of the number of steps to estimate UDE when estimated with $\geq 4$ steps is minimal. While UDE can be calculated with only 2 or 3 steps, this is not sufficient to be able to separate different methods. We see that 4 steps is sufficient to get reasonable estimates, deviating $\pm 0.05$ from the best estimate with 10 steps. Additionally, we see that the noise between runs is minimal, suggesting that they are not required to be able to get good estimates.

\usepgfplotslibrary{groupplots} %
\usetikzlibrary{pgfplots.groupplots} %
\begin{figure}[t]
\centering

\begin{tikzpicture}
\begin{axis}[
    footnotesize,
    height=7cm,
    width=15cm,
    xtick distance=1,
    xmin=2,
    xmax=10,
    xlabel=Steps to estimate UDE,
    ylabel=Estimated UDE,
    ytick distance=0.05,
]

    \pgfplotstableread[col sep=comma]{./data/robustness_it_variants.csv}\rawdata
    
    \pgfplotstablesort[sort key={step}]{\sorted}{\rawdata} %

    \addplot[name path=it, c0] 
        table[x={step}, y={Information Theoretic}]{\sorted};
    
    \addplot[name path=it_upper, draw=none]
        table[x=step,
              y expr=\thisrow{Information Theoretic} + 2* \thisrow{Information Theoretic-std}/ sqrt(3)]
              {\sorted};

    \addplot[name path=it_lower, draw=none]
        table[x=step,
              y expr=\thisrow{Information Theoretic} - 2* \thisrow{Information Theoretic-std}/ sqrt(3)]
              {\sorted};

    \addplot[c0!30, fill opacity=0.3]
        fill between[of=it_upper and it_lower];

    \addplot[name path=kldiv, c1] 
        table[x={step}, y={KL Divergence}]{\sorted};
    
    \addplot[name path=KL Divergence_upper, draw=none]
        table[x=step,
              y expr=\thisrow{KL Divergence} + 2* \thisrow{KL Divergence-std}/ sqrt(3)]
              {\sorted};

    \addplot[name path=KL Divergence_lower, draw=none]
        table[x=step,
              y expr=\thisrow{KL Divergence} - 2* \thisrow{KL Divergence-std} / sqrt(3)]
              {\sorted};

    \addplot[c1!30, fill opacity=0.3]
        fill between[of=KL Divergence_upper and KL Divergence_lower];

    \addplot[name path=biasvariance, c2] 
        table[x={step}, y={Bias-Variance}]{\sorted};
    
    \addplot[name path=Bias-Variance_upper, draw=none]
        table[x=step,
              y expr=\thisrow{Bias-Variance} + 2* \thisrow{Bias-Variance-std} / sqrt(3)]
              {\sorted};

    \addplot[name path=Bias-Variance_lower, draw=none]
        table[x=step,
              y expr=\thisrow{Bias-Variance} - 2* \thisrow{Bias-Variance-std} / sqrt(3)]
              {\sorted};

    \addplot[c2!30, fill opacity=0.3]
        fill between[of=Bias-Variance_upper and Bias-Variance_lower];

   \end{axis}

\end{tikzpicture}

    \caption{Change of estimated Uncertainty Disentanglement Error for \textcolor{c0}{Information Theoretic measures}, \textcolor{c1}{Pairwise KL-Divergence}, and \textcolor{c2}{Bias-Variance measures}, when varying the number of steps in the experiments. Shaded areas indicate two standard errors. Overall, the impact of the number of steps is fairly small when the number of steps $\geq 4$, giving an acceptable method to reduce the computational cost of calculating UDE. }
    \label{fig:robustness_it_alternatives}

\end{figure}

From this, we can conclude that for estimate UDE with ImageNet-1k it is sufficient to calculate it using label noises at $\{0\%, 25\%, 50\%, 75\%\}$ and dataset sizes at $\{25\%, 50\%, 75\%, 100\%\}$, using only a single run. We use a Deep Ensemble with 3 ResNext50 models \cite{xie2017aggregated} trained from random initialization. The results are reported in Table \ref{tab:ImageNetResults_maintext} in the main body of the paper.

These results shows that IT disentanglement performs substantially better on orthogonality, because the correlation between $u_e$ and $U_a$ is much smaller than for Pairwise KL-Divergence or Bias-Variance disentangling. Still, the estimates of $u_a$ are not orthogonal to $U_e$.

\FloatBarrier

\section{Setting Parameters for Uncertainty Disentanglement Error}\label{sec:ude_hyperparameters}

Uncertainty Disentanglement Error can be computed for any model on any dataset if the disentanglement method makes estimates of aleatoric or epistemic uncertainty. It can be computed using different levels of precision for different compute costs. We documented the parameters used in this study, but in this appendix we will also give advice for considerations when applying UDE to different datasets.

A general consideration is the number of repetitions for computing the UDE. In the present study, for each method, model and dataset the UDE was computed 5 times with different seeds for subsampling, noise, and model training. In the results for larger datasets (CIFAR10, FashionMNIST) we found minimal variance between runs, whereas for smaller datasets (BCI, Wine) variance was larger. Based on this observation, we find that for large datasets such as ImageNet-1k it is acceptable to estimate UDE with only a single run. For small datasets where calculating UDE is relatively inexpensive, doing 10 runs for robust results is advisable. 

\subsection{Decreasing Dataset experiment}

For the decreasing dataset experiment a number of dataset percentages needs to be set to establish the correlation between accuracy and uncertainties. The aim is to capture the curve of accuracy seen in Figure 1a. Since the relation between dataset size and accuracy follows a power law, if no substantial change in accuracy is observed, it may help to set the dataset percentage in a log-curve. 

To properly estimate the UDE 10 samples are recommended. In Appendix \ref{sec:imagenet_robustness} we show that similar results are achieved with fewer samples on CIFAR-10. Such an analysis may be used to justify setting fewer samples on a large dataset. 
An important sanity check for the decreasing dataset experiment is to validate that the accuracy substantially changes over the size of the dataset explored. If the accuracy remains mostly constant, the UDE should be calculated over different (smaller) dataset sizes. 

The number of epochs should be kept proportional to the number of samples to make sure the number of batches stays consistent. It is possible to set a high number of epochs for all and use early-stopping to determine the number of epochs. In this case the validation data for early stopping cannot come from the training (sub)set, because smaller validation datasets may add noise to the early stopping mechanism and cause the model to stop too early. Early stopping was not used in the ImageNet-1k experiments reported in Appendix \ref{sec:imagenet_robustness}, because it caused underfitting. 

\subsection{Label Noise experiment}
For the label noise experiment a similar number of 5-10 samples may be used. Unlike the decreasing dataset experiment, the relationship between the label noise and the accuracy is fairly linear, so noise percentages can always be linearly spaced.

\section{Failure Example Information Theoretic}\label{sec:it_triangles}

\begin{figure}[t]
    \begin{subfigure}{.4\linewidth}
      \includegraphics[width=\linewidth]{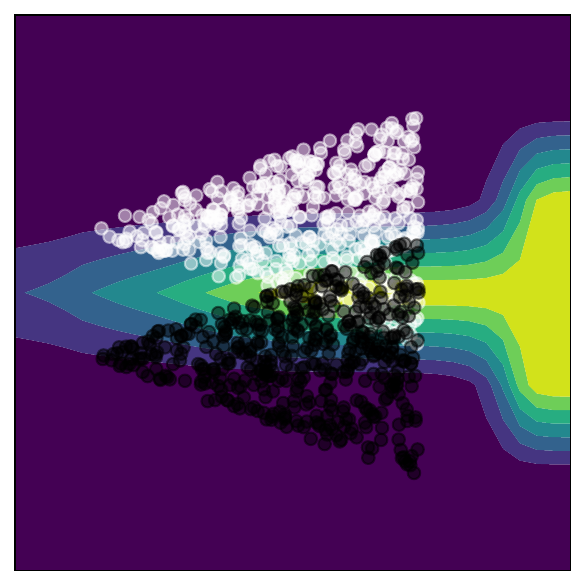}
      \caption{Aleatoric}\label{fig:it_triangles_aleatoric}
    \end{subfigure}%
    \hfill
    \begin{subfigure}{.4\linewidth}
      \includegraphics[width=\linewidth]{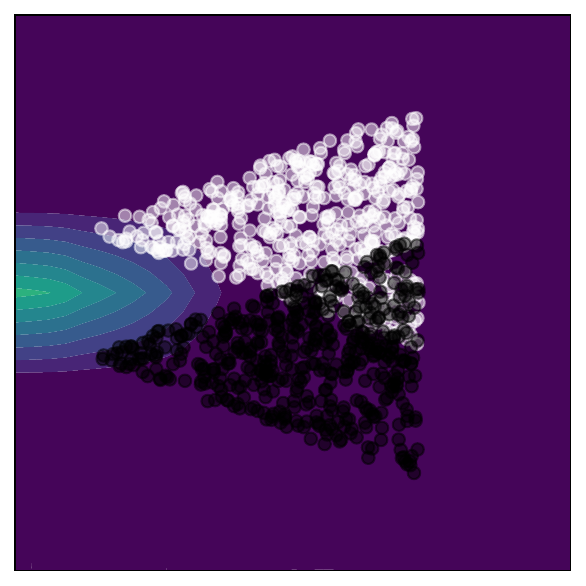}
      \caption{Epistemic}\label{fig:it_triangles_epistemic}
    \end{subfigure}
    \caption{Failure example for Information Theoretic Disentangling. The white and black dots represent artificial training samples. Bright colours in the background represent high uncertainty. Figure \ref{fig:it_triangles_aleatoric} shows an increase in AU further to the right where the classes have more overlap, which is correct. Figure \ref{fig:it_triangles_epistemic} shows EU on the left of the training data but not on the right, which is incorrect. %
    }
    \label{fig:it_triangles}
\end{figure}

Figure \ref{fig:it_triangles} demonstrates a failure case for Information Theoretic disentangling. This example was made with an artificial distribution, designed to have low aleatoric uncertainty on the left side of the feature space, and high aleatoric uncertainty on the right. This effect was implement by generating samples on a triangle-shaped distribution. 

The uncertainty estimations were made with a small Multi-Layer Perceptron with shape $2 \times 32 \times 32 \times 2$, with MC-Dropout applied to both of the hidden layers. Aleatoric and epistemic uncertainty were subsequently estimated using the Information Theoretic disentangling formulation from Equation \ref{eq:it_disentanling}. The results show that under high aleatoric uncertainty (on the right) the epistemic uncertainty is underestimated. This makes Figure \ref{fig:it_triangles} a visualization when the additivity assumption described in \citet{wimmer2023quantifying} results in erroneous interactions.While this is only a demonstration using MC-Dropout, other BNN approxations using IT disentangling show similar behaviour.

\section{Underfitting on Two Moons} \label{sec:underfitting-two-moons}

\begin{figure}[h]
    \centering
    \includegraphics[width=0.9\linewidth]{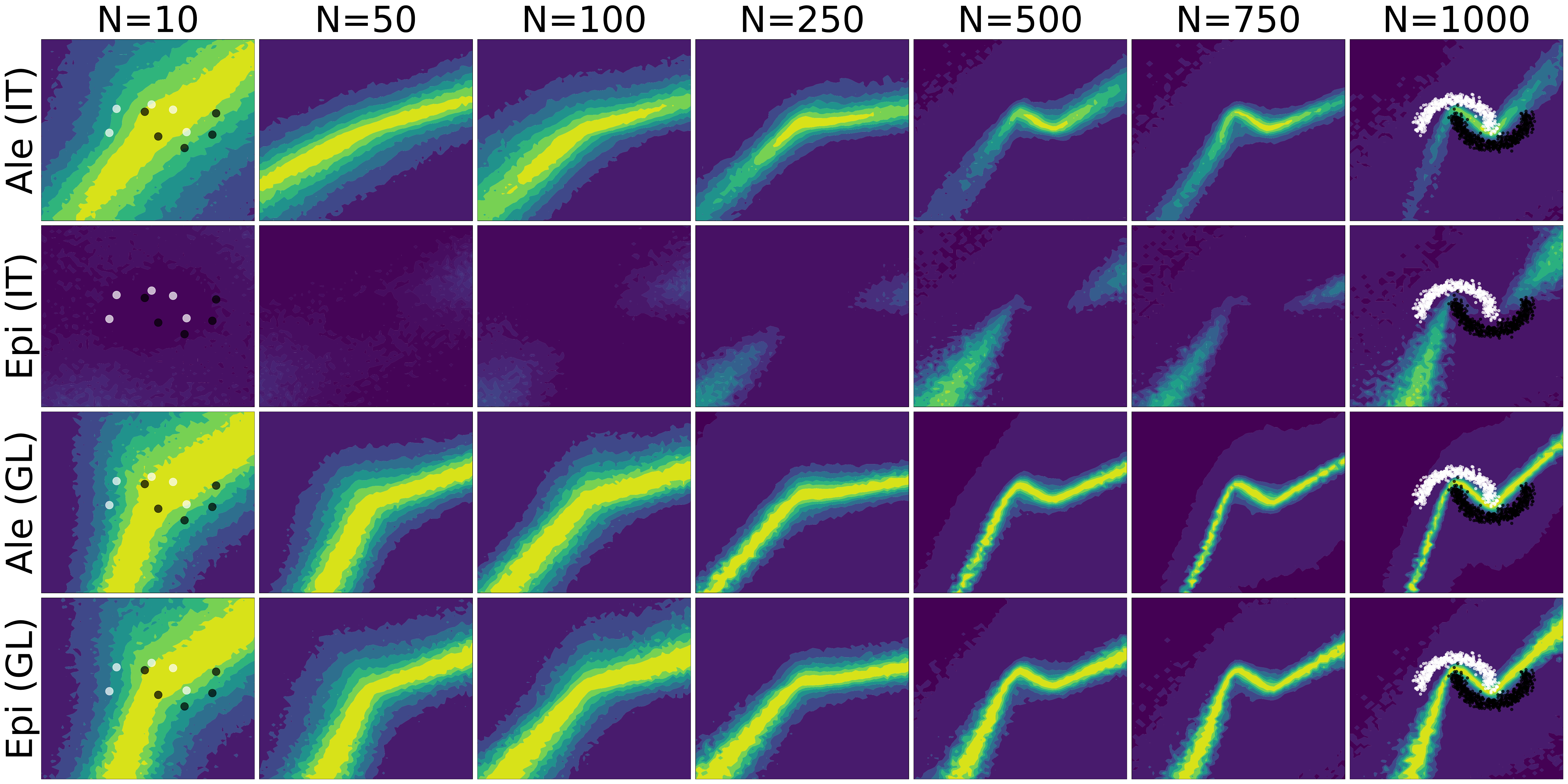}
    \caption{Aleatoric and epistemic uncertainty with changing dataset sizes for the Two Moons dataset with MC-Dropout where each model is trained for 100 epochs. For small datasets (left size) the decision boundary becomes simpler, and the models predicted more AU. On small datasets the models are under-fitting.}
    \label{fig:two-moons-dataset-size-underfitting}
\end{figure}

Figure \ref{fig:two-moons-dataset-size-underfitting} shows a similar setup to Figure \ref{fig:two-moons-dataset-size}, but the amount of epochs is kept at 100. Here we see that for fewer training samples the learned decision boundary is much simpler, so that higher AU is predicted. This shows that an underfitting model can misrepresent EU as AU. This problem exists regardless of the method for disentangling.

Based on this we decided to make the number of epochs inversely proportional to the training dataset size, so we wouldn't falsely estimate too high AU due to underfitting.

\section{Detailed Results}\label{sec:detailed-results}

\begin{table}[h]
\centering
\captionof{table}{Pearson's correlation coefficient $\rho(\cdot, \cdot)$ between $-acc$, representing the true epistemic uncertainty $U_e$ and aleatoric $u_a$ and epistemic $u_e$ uncertainty estimates in the Decreasing Dataset experiment. Ideally, epistemic uncertainty $u_e$ should correlate with $U_e$ ($\uparrow$), while aleatoric uncertainty estimates $u_a$ should remain uncorrelated ($\rightarrow0$).}
\label{tab:dataset_size_correlations}
\begin{tabular}{>{\columncolor[HTML]{EFEFEF}}lcccc}
\toprule
\textbf{Dataset / Method} & \textbf{GL $\rho(u_a, U_e) \rightarrow0$} & \textbf{GL $\rho(u_e, U_e) \uparrow$} & \textbf{IT $\rho(u_a, U_e) \rightarrow0$} & \textbf{IT $\rho(u_e, U_e) \uparrow$} \\
\midrule
\multicolumn{5}{l}{\textit{CIFAR10} \citep{krizhevsky2009learning}} \\
MC-Dropout       & $-0.852$ & $0.13$ & $-0.876$ & \cellcolor{green!25}$0.994$ \\
MC-DropConnect   & $-0.695$ & $0.778$ & $-0.728$ & \cellcolor{yellow!25}$0.917$ \\
Flipout          & \cellcolor{yellow!25}$0.453$ & $0.848$ & $0.130$ & $0.899$ \\
Deep Ensembles   & $-0.900$ & $0.106$ & $-0.902$ & \cellcolor{green!25}$0.958$ \\
\midrule
\multicolumn{5}{l}{\textit{Fashion MNIST} \citep{xiao2017fashion}} \\
MC-Dropout       & $-0.791$ & $-0.106$ & $-0.729$ & \cellcolor{green!25}$0.997$ \\
MC-DropConnect   & \cellcolor{yellow!25}$0.225$ & $0.765$ & \cellcolor{yellow!25}$0.416$ & $0.939$ \\
Flipout          & $-0.919$ & \cellcolor{green!25}$0.976$ & $-0.949$ & $0.990$ \\
Deep Ensembles   & $-0.669$ & \cellcolor{yellow!25}$0.967$ & $-0.763$ & \cellcolor{green!25}$0.996$ \\
\midrule
\multicolumn{5}{l}{\textit{Wine} \citep{misc_wine_109}} \\
MC-Dropout       & $-0.889$ & $-0.746$ & $-0.875$ & $0.728$ \\
MC-DropConnect   & \cellcolor{yellow!25}$-0.178$ & \cellcolor{yellow!25}$0.024$ & $-0.598$ & $0.677$ \\
Flipout          & $-0.524$ & \cellcolor{green!25}$0.998$ & \cellcolor{yellow!25}$0.122$ & \cellcolor{green!25}$0.995$ \\
Deep Ensembles   & $-0.426$ & $0.691$ & $-0.842$ & $0.928$ \\
\midrule
\multicolumn{5}{l}{\textit{BCI} \citep{brunner2008bci}} \\
MC-Dropout       & $-0.944$ & $-0.873$ & $-0.921$ & $0.787$ \\
MC-DropConnect   & \cellcolor{yellow!25}$0.697$ & $0.194$ & $-0.248$ & \cellcolor{yellow!25}$-0.893$ \\
Flipout          & $-0.961$ & $0.718$ & $-0.918$ & $0.879$ \\
Deep Ensembles   & $-0.964$ & $-0.403$ & $-0.965$ & \cellcolor{green!25}$0.976$ \\
\midrule
\multicolumn{5}{l}{\textbf{Average} (absolute value for $\rho(u_a, U_e)$)} \\
MC-Dropout       & $0.869$ & $-0.399$ & $0.850$ & $0.877$ \\
MC-DropConnect   & \cellcolor{yellow!25}$0.449$ & $0.440$ & \cellcolor{green!25}$0.498$ & $0.410$ \\
Flipout          & $0.714$ & $0.885$ & $0.530$ & \cellcolor{yellow!25}$0.941$ \\
Deep Ensembles   & $0.740$ & $0.340$ & $0.868$ & \cellcolor{green!25}$0.964$ \\
\bottomrule
\end{tabular}
\end{table}

\begin{table}[t]
\centering
\caption{Pearson's correlation coefficient $\rho(\cdot, \cdot)$ of $-acc$, representing the true aleatoric uncertainty $U_a$ and epistemic $u_e$ and aleatoric $u_a$ uncertainty estimates under Label Noise. Aleatoric estimate $u_a$ should ideally correlate with $U_a$ ($\rightarrow 1$), epistemic estimates $u_e$ should remain uncorrelated ($\rightarrow 0$).}
\label{tab:label_noise_pcc}
\begin{tabular}{>{\columncolor[HTML]{EFEFEF}}lcccc}
\toprule
\textbf{Dataset / Method} & \textbf{GL $\rho(u_a, U_a) \uparrow$} & \textbf{GL $\rho(u_e, U_a) \rightarrow 0$} & \textbf{IT $\rho(u_a, U_a) \uparrow$} & \textbf{IT $\rho(u_e, U_a) \rightarrow 0$} \\
\midrule
\multicolumn{5}{l}{\textit{CIFAR10} \citep{krizhevsky2009learning}} \\
MC-Dropout       & $0.962$ & $0.975$ & $0.933$ & \cellcolor{green!25}$0.266$ \\
MC-DropConnect   & \cellcolor{green!25}$0.996$ & $0.996$ & \cellcolor{yellow!25}$0.981$ & $0.486$ \\
Flipout          & $0.751$ & $0.725$ & $0.585$ & $0.484$ \\
Deep Ensembles   & $0.974$ & $0.996$ & $0.927$ & \cellcolor{yellow!25}$0.312$ \\
\midrule
\multicolumn{5}{l}{\textit{Fashion MNIST} \citep{xiao2017fashion}} \\
MC-Dropout       & $0.990$ & $0.991$ & $0.963$ & \cellcolor{green!25}$0.258$ \\
MC-DropConnect   & \cellcolor{green!25}$0.997$ & $0.997$ & \cellcolor{green!25}$0.995$ & $0.649$ \\
Flipout          & $0.958$ & $0.931$ & $0.932$ & $-0.604$ \\
Deep Ensembles   & $0.982$ & $0.994$ & $0.991$ & \cellcolor{yellow!25}$0.561$ \\
\midrule
\multicolumn{5}{l}{\textit{UCI Wine} \citep{misc_wine_109}} \\
MC-Dropout       & $0.953$ & $0.965$ & $0.943$ & $-0.873$ \\
MC-DropConnect   & $0.961$ & $0.976$ & $0.945$ & \cellcolor{green!25}$0.614$ \\
Flipout          & \cellcolor{yellow!25}$0.983$ & $0.988$ & \cellcolor{green!25}$0.984$ & $0.916$ \\
Deep Ensembles   & $0.968$ & $0.972$ & $0.960$ & \cellcolor{yellow!25}$0.700$ \\
\midrule
\multicolumn{5}{l}{\textit{BCI} \citep{brunner2008bci}} \\
MC-Dropout       & $0.975$ & $0.972$ & $0.976$ & $0.947$ \\
MC-DropConnect   & $0.968$ & $0.977$ & $0.962$ & \cellcolor{green!25}$0.915$ \\
Flipout          & $0.979$ & $0.985$ & $0.953$ & \cellcolor{yellow!25}$0.926$ \\
Deep Ensembles   & \cellcolor{yellow!25}$0.991$ & $0.990$ & \cellcolor{green!25}$0.996$ & $0.932$ \\
\midrule
\multicolumn{5}{l}{\textbf{Average} (absolute value for EU)} \\
MC-Dropout       & $0.970$ & $0.976$ & $0.954$ & \cellcolor{green!25}$0.586$ \\
MC-DropConnect   & \cellcolor{green!25}$0.981$ & $0.987$ & $0.971$ & $0.666$ \\
Flipout          & $0.918$ & $0.907$ & $0.864$ & $0.733$ \\
Deep Ensembles   & \cellcolor{yellow!25}$0.979$ & $0.988$ & $0.969$ & \cellcolor{yellow!25}$0.626$ \\
\bottomrule
\end{tabular}
\end{table}
\FloatBarrier

\section{Full Visualizations for Uncertainty Disentanglement Error} \label{sec:full_results}

\subsection{Decreasing dataset}

Figure \ref{fig:decreasing_dataset_Fashion_MNIST} shows that the results on decreasing dataset are very similar between Fashion MNIST and CIFAR10. The main difference is that the overall accuracy is higher, and the overall uncertainty is lower. When we look at the results for the Wine dataset in Figure \ref{fig:decreasing_dataset_Wine} we see much more noise due to the small dataset, but the overall pattern is still the same. Flipout and Deep Ensembles show a clear decrease in EU, while AU increases. For MC-Dropout and MC-Dropconnect the epistemic uncertainty does not always clearly decrease.

The same setup of the three experiments is repeated with a CNN on the Fashion MNIST dataset \citep{xiao2017fashion}. Figure \ref{fig:decreasing_dataset_Fashion_MNIST} shows the results of the Dataset Size experiment on Fashion MNIST. The accuracy is higher than for the CIFAR-10 dataset, and the uncertainty is lower, but the patterns overall are very similar for the different UQ methods and the different disentanglement approaches. The results for the Wine dataset in Figure \ref{fig:decreasing_dataset_Wine} show more noise, possibly because there is not such a big decrease in accuracy. We see that on this dataset Flipout gives exceptionally good uncertainty estimation, as also reflected in the summary statistics in Table \ref{tab:dataset_size_correlations}. 

In Figure \ref{fig:decreasing_dataset_bci} we show the results on the BCI dataset. The results for IT with MC-Dropconnect show that because MC-Dropconnect fails to predict a change in EU, the AU also stays consistent. This shows that interactions of AU and EU are not inherent in the data, but a consequence of how they are estimated.

\input{tikzfigures/dataset_size_fashion_mnist.tex}

\input{tikzfigures/dataset_size_wine.tex}

\input{tikzfigures/dataset_size_bci}

\subsection{Label Noise}
Figure \ref{fig:label_noise_Fasion_MNIST} shows again that there are almost no difference between the results for Fashion MNIST and CIFAR10 for the Label Noise experiment. The different BNNs cause much larger differences in estimated uncertainties than the different datasets. We see that the Wine dataset gives the same result, but with more noise in Figure \ref{fig:decreasing_dataset_Wine}. The BCI results in Figure \ref{fig:label_noise_bci} do not show such a strong increase in AU because the uncertainty at 0\% shuffled is already high. The increase is consistent, which is also reflected by the summary statistics in Table \ref{tab:label_noise_pcc}.

\input{tikzfigures/label_noise_fashion_mnist.tex}

\input{tikzfigures/label_noise_wine.tex}

\input{tikzfigures/label_noise_bci.tex}

\FloatBarrier
\newpage

%
%

%

%

%

%
%

%
%
%
%
%
%
%

%

%

%

%

%

%
%
%

%

%

%

%
%
%
%
%
%
    
%
%
%
%
%
    
%
%
%
%
%
    
%
%
%

%

%
%
%
%
%
%
%
%
%
%
%
%
%
%
%
%
%
%
%
%
%
%

%

\FloatBarrier

\end{document}